%% file: main.tex
\documentclass[runningheads]{llncs}
\usepackage{eccv}

\usepackage{eccvabbrv}

\usepackage{graphicx}
\usepackage{booktabs}

\usepackage[accsupp]{axessibility}  

\usepackage{wrapfig}

\usepackage{hyperref}

\usepackage{orcidlink}

\begin{document}

\title{Regularizing Dynamic Radiance Fields with Kinematic Fields}


\author{Woobin Im\inst{1}\thanks{This work was done during an internship at NAVER Cloud.} \and
Geonho Cha \inst{2} \and
Sebin Lee \inst{1} \and
Jumin Lee \inst{1} \and
Juhyeong Seon \inst{1} \and
Dongyoon Wee \inst{2} \and
Sung-Eui Yoon \inst{1}
}

\authorrunning{W. Im et al.}

\institute{KAIST\\
\email{\{iwbn,seb.lee,jmlee,munuwazzi,sungeui\}@kaist.ac.kr}
\and
NAVER Cloud\\
\email{\{geonho.cha,dongyoon.wee\}@navercorp.com}}

\maketitle

\input{sec/0_abstract}    
\input{sec/1_intro}
\input{sec/2_related_work}
\input{sec/3_method}
\input{sec/4_experiment}

\input{sec/5_conclusion}

{
\makeatletter
\renewcommand\paragraph{\@startsection{paragraph}{4}{\z@}%
{3.25ex \@plus1ex \@minus.2ex}%
{-1em}%
{\normalfont\normalsize\bfseries}}
\makeatother
\paragraph{Acknowledgment.}
This project was partly supported by the NAVER Cloud Corporation. 
Additionally, this work received support from the Institute of Information \& communications Technology Planning \& Evaluation~(IITP)
grant (RS-2023-00237965) and the National Research Foundation of Korea~(NRF) grant (No. RS-2023-00208506(2024)), funded by the Korea government~(MSIT). Sung-Eui Yoon is a corresponding author.
}

\bibliographystyle{splncs04}
\bibliography{main}
\clearpage

\input{sec/6_supplementary}
\end{document}

%% file: sec/0_abstract.tex
\begin{abstract}
This paper presents a novel approach for reconstructing dynamic radiance fields from monocular videos.
We integrate kinematics with dynamic radiance fields, bridging the gap between the sparse nature of monocular videos and the real-world physics.
Our method introduces the kinematic field, 
capturing motion through kinematic quantities: velocity, acceleration, and jerk.
The kinematic field is jointly learned with the dynamic radiance field by minimizing the photometric loss without motion ground truth.
We further augment our method
with physics-driven regularizers grounded in kinematics.
We propose physics-driven regularizers that ensure the physical validity of predicted kinematic quantities, including advective acceleration and jerk. 
Additionally, we control the motion trajectory based on rigidity equations formed with the predicted kinematic quantities.
In experiments, our method outperforms the state-of-the-arts by capturing physical motion patterns within challenging real-world monocular videos.

\keywords{Dynamic radiance fields \and Kinematic fields \and 3D video}
\end{abstract}

%% file: sec/1_intro.tex
\section{Introduction}

Recognizing motion is crucial to how we visually perceive the world~\cite{hess1989motion}, and video is the medium that carries the motion to viewers in a remote place or at a different time.
Through the lens of video, we watch and enjoy movies, user-generated contents, and remote video chats.
While videos depict motion in a 2D format, transforming them into three-dimensional experiences can elevate our perception of the content to co-presence via applications such as teleportation~\cite{orts2016holoportation}.

Dynamic radiance fields~\cite{pumarola2021d,li2022neural,li2023dynibar,cao2023hexplane,gao2022monocular,fang2022fast} have been studied as a promising representation for reconstructing 3D dynamic scenes; it enables realistic novel view and time synthesis,
which can upgrade 2D videos into 3D videos.
Since multi-view cameras can give more information about the geometry in a scene, existing works have utilized camera rigs with dozens of cameras to capture multi-view video datasets~\cite{li2022neural,attal2023hyperreel,cao2023hexplane}. 
Although high-quality multi-view videos result in undeniably better quality than monocular videos do, the multi-view capturing devices and environments are prohibitive in most cases.

Thus, there have been efforts to reconstruct dynamic radiance fields from monocular videos~\cite{li2021neural,Gao_2021_ICCV,Liu_2023_CVPR}. 
Compared to multi-view videos, monocular videos provide sparser information, which makes the reconstruction more challenging. 
For instance, there is scale ambiguity in monocular video, making it difficult to accurately locate an object within a space~\cite{wang2023flow}.
Thus, dynamic scene reconstruction from monocular videos has been considered an ill-posed problem.

In this paper, we incorporate kinematics to fill the gap between sparse monocular videos and real-world physics, potentially enhancing novel view and time synthesis.
Our core idea is that motion within the real-world is governed by the principles of kinematics in physics, implying that representations of the world, such as videos, are inherently anchored in kinematics.
Building on this foundation, we generalize conventional displacement fields -- often called scene-flow fields~\cite{li2021neural} -- to kinematic fields.
In prior work, motion in radiance fields has been controlled via the regularization in the frequency domain~\cite{wang2021neural,ramasinghe2023bali}, displacement fields~\cite{li2021neural}, and deformation fields~\cite{park2021nerfies}.

Different from these approaches, 
our kinematic field regulates motion in radiance fields using kinematics;
it innately creates continuous trajectories and enables us to impose regularization based on kinematics.
We integrate physics-based regularizers to make the kinematic field physically plausible and thus improve the dynamic radiance field accordingly. 
For instance, the rigidity regularizer based on the first and the second invariant of the strain rate is applied on the kinematic field, suppressing non-rigid motion in the field. 
We quantitatively and qualitatively 
validate the proposed regularizers on the NVIDIA Dynamic Scenes  (NDVS) dataset,
showing the effectiveness of our method over the existing models.
Lastly, the kinematic field uncovers the kinematics within a scene without a ground truth; it estimates not only point-wise 3D displacement within a scene, but also underlying velocity, acceleration, and jerk.

%% file: sec/2_related_work.tex
\section{Related Work}
\noindent \textbf{Neural radiance fields.}
In recent years, neural radiance fields~\cite{mildenhall2020nerf,barron2021mip,verbin2022ref}
have emerged as a promising approach for learning an implicit neural representation of a 3D scene. 
Combined with the differentiable volume rendering technique, it has enabled realistic novel-view synthesis~\cite{mildenhall2020nerf,barron2021mip,verbin2022ref,zhang2020nerf++}.

In essence, a radiance field $\mathcal{F}_\text{ST}$ maps a 5D vector $(x,y,z,\theta, \phi)$ to a 4D output vector $(\mathbf{c}_s, \sigma_s)$:
\begin{equation}
    \mathcal{F}_\text{ST}:
    (x,y,z,\theta,\phi) \rightarrow (\mathbf{c}_s,\sigma_s),
    \label{eq:st_radiance}
\end{equation}
where the view-dependent color $\mathbf{c}_s = (r,g,b)$ and density $\sigma_s$ are learned in relation to the spatial position $(x, y, z)$ and the viewing direction $(\theta, \phi)$.
With the volumetric representation learned via an iterative optimization, volume rendering techniques~\cite{kajiya1984volume, max2005local, mildenhall2020nerf} are employed to aggregate colors and densities along a cast ray. Consequently, this leads to realistic novel-view synthesis
from the inferred radiance, after being trained on dozens of images.

\noindent \textbf{Dynamic radiance fields.}
Radiance fields also facilitate the novel-view synthesis of scenes with moving objects~\cite{li2022neural,park2021nerfies,park2021hypernerf,attal2023hyperreel,cao2023hexplane}.
While this is more challenging than the static case, dynamic radiance fields incorporating motion allow for the rendering of a frame at a desired time and viewpoint. As a result, it can be utilized to make visual effects such as bullet time, stabilized, dolly zoomed, and stereo videos.
Commonly, the time-dependent radiance fields are referred to as \textit{dynamic radiance fields}, a term we will adopt throughout this paper.

In dynamic radiance fields, radiance is dependent on a varying time $t$. This modifies Eq.~\ref{eq:st_radiance} by introducing a time-dependent term, resulting in:
\begin{equation}
\mathcal{F}_\text{DY}:
(x,y,z,t, \theta,\phi) \rightarrow (\mathbf{c}_d,\sigma_d),
\label{eq:dy_radiance}
\end{equation}
where the color and density are conditioned on time $t$.
In this formulation, moving geometries of objects (\textit{e.g.}, translation and deformation) and changing radiance (\textit{e.g.},  moving shadows)
can be learned via optimization.

Learning a dynamic representation presents inherent challenges due to the introduction of temporal dimension. To optimize this representation, a high-quality multi-view dataset can be employed to obtain fine rendering results~\cite{li2022neural,attal2023hyperreel,cao2023hexplane}. 
Meanwhile, realistically captured imagery of a scene tends to be non-synchronized and is likely to be obtained with a limited number of cameras and restricted displacements; this often results in a monocular setup~\cite{gao2022monocular,park2021nerfies,park2021hypernerf,li2021neural}. 
Previous researches have delved into deformable models for monocular videos~\cite{tretschk2021non,park2021nerfies,park2021hypernerf,Liu_2023_CVPR}. In these methods, a deformation field maps a spatio-temporal coordinate to a reference canonical space. 
While the canonical space aids in maintaining radiance consistency across global time frames, it tends to exhibit reduced effectiveness in generic monocular videos, where new objects might appear within a video~\cite{ramasinghe2023bali}. 
In this study, we embrace the concept of scene flow fields~\cite{li2021neural}, which is constructed without canonical space and deformation networks.

\noindent \textbf{Regularization with kinematic priors.}
Monocular videos present challenges in reconstruction due to the constrained viewpoints of moving objects. 
To address this, integrating auxiliary flow fields has proven to be effective~\cite{li2021neural,Gao_2021_ICCV,du2021neural,li2023dynibar}. Existing studies employ either scene flow fields~\cite{li2021neural,Gao_2021_ICCV,li2023dynibar} or velocity fields~\cite{du2021neural} to capture spatial displacements 
between consecutive frames.
Since the scene flows can give spatial correspondences of the dynamic radiance field, it can be used to promote photometric consistency between the renderings of a given ray and its time-deformed counterpart. 

Physical priors incorporated with scene flow fields play a pivotal role in regulating the radiance field. 
Regularization of motion has been applied to the dynamic field, bridging the gap between the sparse visual data and plausible real-world motion.
For instance, a smoothness regularizer ensures that the displacements within close proximity are similar to each other,
and kinetic energy regularizer promotes consistency between forward and backward flows, resulting in smoother displacement fields~\cite{li2021neural}.
More deeply integrated with physics, the elastic energy regularizer minimizes the deformation between the canonical space and the deformed space~\cite{park2021nerfies}.
In fluid reconstruction works~\cite{chu2022physics,baieri2023fluid},
velocity fields grounded in physics have been shown to be 
critical in accurate 4D reconstruction. 
For instance, Navier-Stokes and transport equations can be integrated in reconstructing smoke~\cite{chu2022physics} and fluid surfaces~\cite{baieri2023fluid} with dynamic radiance fields
to reconstruct more physically plausible structures.

We integrate kinematics to reconstruct dynamic radiance fields from generic monocular videos.
We introduce the kinematic field, which captures motion through kinematic quantities such as velocity, acceleration, and jerk.
Different from existing velocity field-based methods~\cite{chu2022physics,du2021neural,wang2023flow},
our method has higher-order terms (\textit{i.e.}, acceleration and jerk) as latent of spatio-temporal correspondences.
Further enhancing our approach, we incorporate physics-driven regularizers based on a transport equation~\cite{chu2022physics} and the strain rate tensor, addressing dynamics of moving particles and rigid motion. 

%% file: sec/3_method.tex
\section{Method}

\begin{figure}[t]
    \centering
    \includegraphics[trim={0cm 7.14cm 5.37cm 0cm},clip,width=0.95\linewidth]{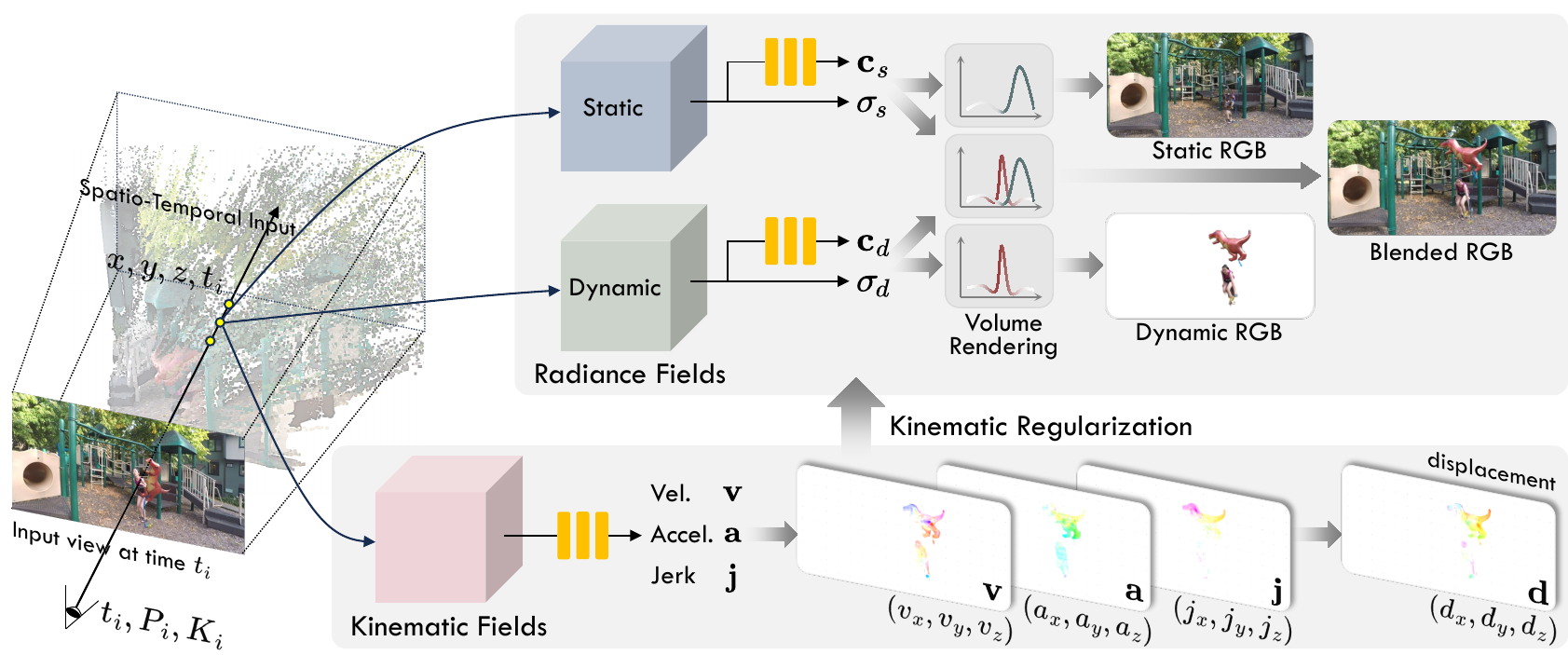}
    \caption{\textbf{Radiance and kinematic fields.}
    This figure summarizes the three fields our method utilizes. The static and dynamic radiance fields are used for rendering, and the kinematic field is used in the training phase, regularizing the dynamic radiance field.
    }
    \label{fig:render}
\end{figure}

\subsection{Overview}

Our system takes as input a video sequence, denoted as ${(I_i, t_i, P_i)}_{i=1}^{N}$,  where $N$ is the number of frames. For each frame, we have 
an image $I_i \in \mathbb{R}^{H \times W \times 3}$, a timestamp $t_i \in \mathbb{R}$, and a precomputed camera pose $P_i \in SE(3)$ and  intrinsic matrix $K_i \in \mathbb{R}^{3 \times 3}$.
From this input, our primary objective is to synthesize an image, denoted as $\hat{I}_k$, corresponding to a novel time and viewpoint $(t_k, P_k, K_k)$. The parameters $t_k$, $P_k$, and $K_k$ are not restricted to those found in the original video input, allowing our system to render images from previously unobserved times and viewpoints.

To achieve this image synthesis for novel times and views, we leverage the radiance-field-based volume rendering~\cite{mildenhall2020nerf}.
Inspired by existing methods~\cite{li2021neural,wu2022d}, we train two distinct radiance fields (Fig.~\ref{fig:render}): a dynamic radiance field, denoted as $\mathcal{F}_\text{DY}$ (Eq.~\ref{eq:dy_radiance}), and a static radiance field, denoted as $\mathcal{F}_\text{ST}$ (Eq.~\ref{eq:st_radiance}). 
To blend the two fields, we utilize a technique from D$^2$-NeRF~\cite{wu2022d}, where rendering weights for dynamic and static samples are derived from the volume densities.

Addressing the challenge of achieving physically plausible spatio-temporal structures, we introduce the novel kinematic field.
The kinematic field accounts for the kinematic quantities (\textit{e.g.}, velocity and acceleration). 
These quantities can be converted into displacement (Eq.~\ref{eq:taylor}),
enabling us to use the photometric consistency loss (Eq.~\ref{eq:photo_loss}) for spatio-temporal regularization, as used in previous work~\cite{li2021neural}.
With these kinematic estimations, we apply physics-based regularization techniques. These techniques involve density transport (Eq.~\ref{eq:transport}) and motion rigidity through the strain rate tensor (Eq.~\ref{eq:strain}). Combining the photometric loss (Eq.~\ref{eq:photo_loss}) with these techniques enhances the reconstruction of radiance and geometry, leading to improved novel time and view synthesis performance 
from underdetermined data such as a monocular video.

\subsection{Dynamic and Static Radiance Fields}

In monocular dynamic radiance field methods, it is common to decompose the radiance fields into dynamic and static fields~\cite{li2021neural,gao2022monocular,Liu_2023_CVPR} to prevent undesirable motion in the static area.
We employ tensor-decomposition-based  representations~\cite{cao2023hexplane,chen2022tensorf} for both dynamic and static fields. Specifically, for the dynamic radiance field $\mathcal{F}_\text{DY}$, we use HexPlane~\cite{cao2023hexplane}, and for the static radiance field $\mathcal{F}_\text{ST}$, we use TensoRF~\cite{chen2022tensorf}. Further details on the tensor decomposition can be found in the supplementary material.

\noindent\textbf{Definition.}
The dynamic and static radiance fields are defined in Eq.~\ref{eq:st_radiance}-\ref{eq:dy_radiance} ($\mathcal{F}_\text{ST}$ and $\mathcal{F}_\text{DY}$).
Here, the inputs correspond to the position $(x, y, z)$, the timestamp $t$, and the viewing direction $(\theta, \phi)$ from the camera origin. Each field yields the color $\mathbf{c}$ and the volume density $\sigma$.
Note that the static field is independent of 
time $t$, ensuring a consistent structure throughout the whole time span.

\begin{figure}[ht]
    \centering
\includegraphics[width=0.8\linewidth,trim={0 0 2.5cm 0},clip]{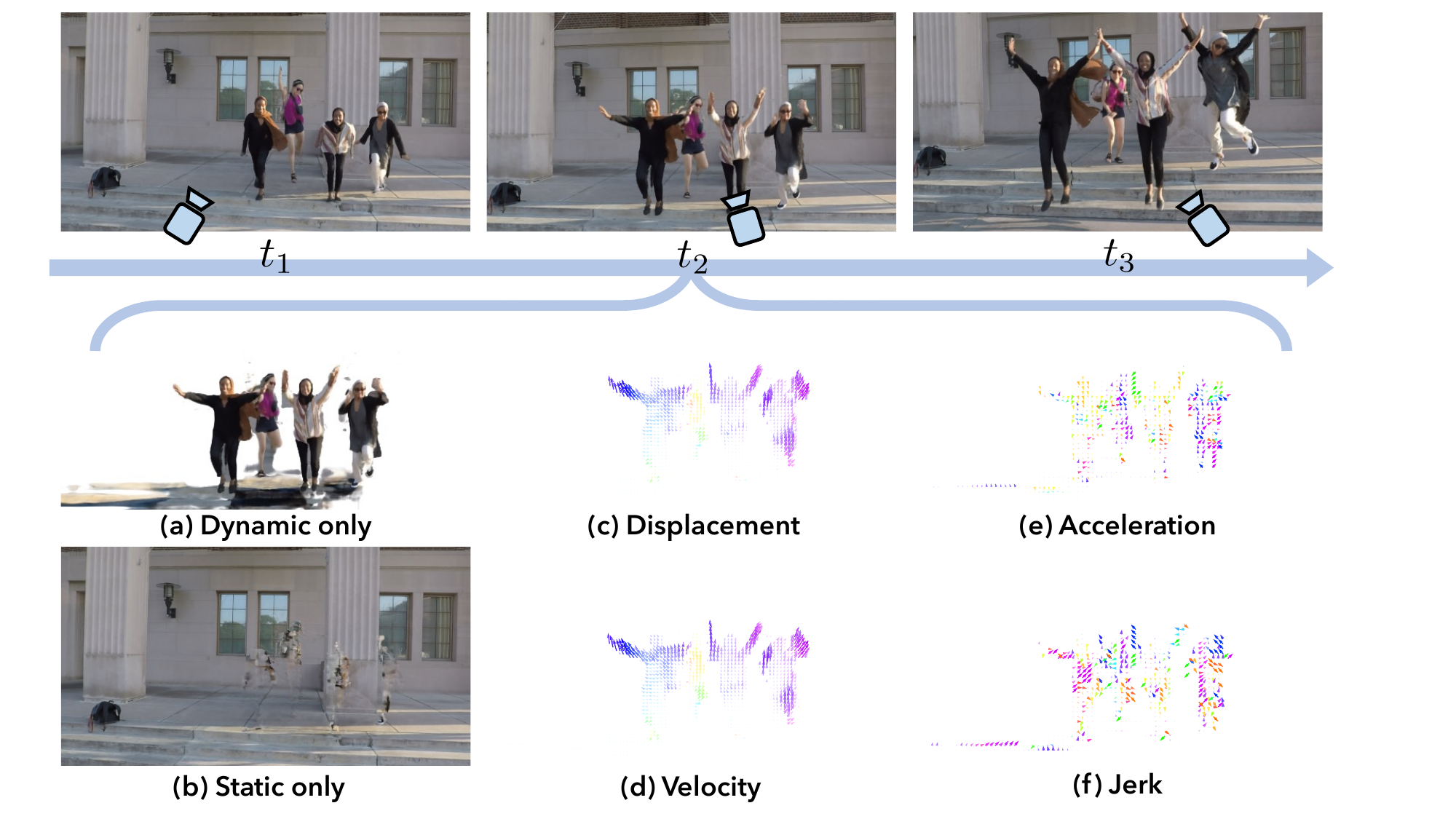}
    \caption{\textbf{Visualization of each predicted component.} The top row presents synthesized views at different times.
    In (a) and (b), the dynamic and static components of the scene at time $t_2$ are depicted separately.
    (b) In the case of an inobservable static area in the entire sequence (\eg, the space behind the jumping people), radiance might not be correct.
    The displacement field (c) can be computed by Taylor approximation (Eq.~\ref{eq:taylor}) with motion fields (d-f): velocity, acceleration, and jerk. Each field is visualized through reprojecting each field to the camera view.
    The standard HSV visualization~\cite{baker2011database} is used to colorize arrows.
    }
    \label{fig:seperation}
\end{figure}

\noindent\textbf{Composite rendering.}
We use a composite rendering method to obtain the final rendering with the two radiance fields.
It is to be noted that individual components (\textit{i.e.}, dynamic-only and static-only) can be rendered using the standard volume rendering technique~\cite{mildenhall2020nerf}.
To combine both fields, we employ an additive composition method in which each field influences the transmittance of a ray~\cite{wu2022d}. Let's consider a camera ray $\mathbf{r}(\tau)=\mathbf{o} + \tau\mathbf{u}$, where $\mathbf{o}$ is the origin and $\mathbf{u}$ is the direction. With a predefined range for the ray distance $[\tau_n, \tau_f]$, the rendering function is formulated as:
\begin{align}
\hat{C}(\mathbf{r}, t_i) = \int_{\tau_n}^{\tau_f} T(\tau) \left( \sigma_s(\tau) \cdot \mathbf{c}_s(\tau) + \sigma_d(\tau, t_i) \cdot \mathbf{c}_d(\tau, t_i) \right) d\tau, \\
\text{where }
T(\tau) = \exp \left( -\int_{\tau_n}^{\tau} (\sigma_s(u) + \sigma_d(u, t_i)) du \right).\label{eq:vol_render}
\end{align}
Here, each notation for the sample point is simplified for concise representation (\textit{e.g.}, $\sigma_s(\tau) = \sigma_s(\mathbf{r}(\tau))$).
We visualize the example of the composite rendering and separate rendering in the first and second rows in Fig.~\ref{fig:seperation}.

\noindent\textbf{Dynamic and static separation.}
To factorize a scene into static and dynamic parts, 
we use a metric indicating the likelihood that a point belongs to the dynamic fields~\cite{wu2022d}.
This metric is given by:
\begin{equation}
    p_d = \frac{\sigma_d}{\sigma_d + \sigma_s},
\end{equation}
where $\sigma_d$ and $\sigma_s$ are density values sampled from a spatio-temporal point.

To facilitate the separation of the radiance fields, we employ the skewed binary entropy loss, as proposed by~\cite{wu2022d}.
The entropy loss is defined as:
\begin{equation}
\mathcal{L}_b = -(p_d^k \cdot \log(p_d^k) + (1 - p_d^k) \cdot \log(1 - p_d^k)),\label{eq:ent}
\end{equation}
where the exponent $k \geq 1$ is the skew factor, controlling a bias toward static explanation.
This entropy-based loss can encourage the radiance fields to have a clearer distinction between the static and dynamic components.

\subsection{Kinematic Field}
The kinematic field has a spatio-temporal structure to facilitate spatial and temporal queries. 
We use the tensor-decomposition-based representation~\cite{cao2023hexplane} 
for the kinematic field. 
As depicted in Fig.~\ref{fig:render},
the kinematic field serves to regulate the radiance field, ensuring that the motion in the radiance field is physically plausible.

\noindent \textbf{Definition.} 
The kinematic field is defined as:
\begin{equation}
\mathcal{F}_\text{K}:
(x,y,z,t) \rightarrow (\mathbf{v}, \mathbf{a}, \mathbf{j}, \cdots),
\end{equation}
where each vector represents velocity, acceleration, and jerk, such that $\mathbf{v}=(v_x, v_y, v_z)$, $\mathbf{a}=(a_x, a_y, a_z)$, and $\mathbf{j}=(j_x, j_y, j_z)$.
Note that we can extend the field into higher-order quantities such as snap (4th), crackle (5th), and so on.

\noindent \textbf{Displacement.}
Using the provided kinematic quantities, we can approximate a particle's displacement at position $(x,y,z)$ and time $t + \Delta t$ via the Taylor approximation, assuming that the
kinematic quantities are consistent within a local temporal range $\Delta t$.
The displacement $d(x,y,z,t,\Delta t)$ can be approximated by:
\begin{equation}
    d(x,y,z,t,\Delta t) \approx \frac{(\Delta t)\mathbf{v}}{1!} 
                  + \frac{(\Delta t)^2\mathbf{a}}{2!}
                  + \frac{(\Delta t)^3\mathbf{j}}{3!} + \cdots
    .\label{eq:taylor}
\end{equation}
Computing the displacement of a point within 
a temporal range allows for computing the photometric consistency loss (Eq.~\ref{eq:photo_loss}) and scene flows from the kinematic field. 

\begin{figure}[t]
    \centering
    \begin{subfigure}[b]{0.49\linewidth}
         \centering
         \begin{subfigure}[b]{0.15\linewidth}
            \includegraphics[width=\textwidth,trim=300 20 150 30,clip]{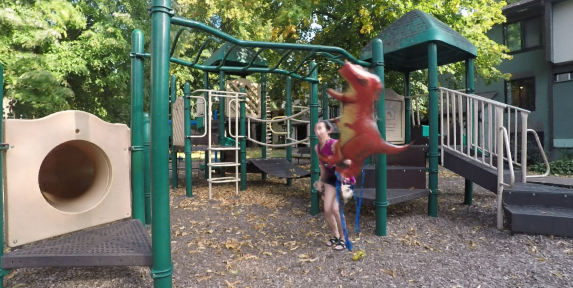}
            \caption*{RGB}
         \end{subfigure}
         \begin{subfigure}[b]{0.15\linewidth}
         \includegraphics[width=\textwidth,trim=300 20 150 30,clip]{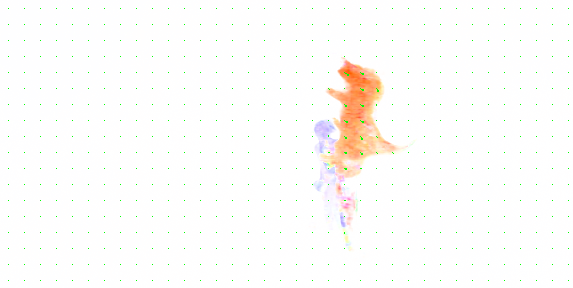}
         \caption*{$\mathbf{v}$}
         \end{subfigure}
         \begin{subfigure}[b]{0.15\linewidth}
         \includegraphics[width=\textwidth,trim=300 20 723 30,clip]{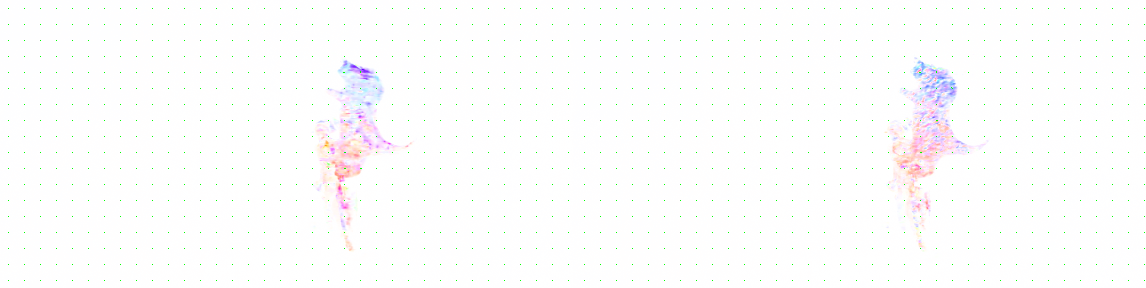}
         \caption*{$\mathbf{a}$}
         \end{subfigure}
         \begin{subfigure}[b]{0.15\linewidth}
         \includegraphics[width=\textwidth,trim=873 20 150 30,clip]{fig/kine_effect/test016_acc_fw.png}
         \caption*{Adv. $\mathbf{a}$}
         \end{subfigure}
         \begin{subfigure}[b]{0.15\linewidth}
         \includegraphics[width=\textwidth,trim=300 20 723 30,clip]{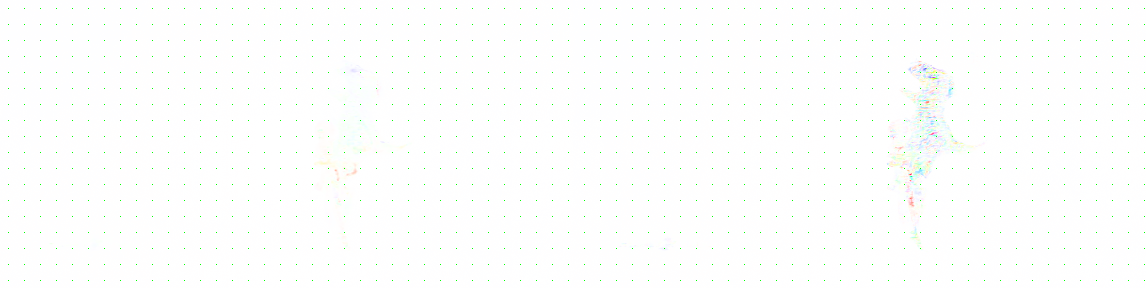}
         \caption*{$\mathbf{j}$}
         \end{subfigure}
         \begin{subfigure}[b]{0.15\linewidth}
         \includegraphics[width=\textwidth,trim=873 20 150 30,clip]{fig/kine_effect/test016_jer_fw-2.png}
         \caption*{Adv. $\mathbf{j}$}
         \end{subfigure}
         
         \caption{Without kinematic regularization}
         \label{fig:kinematic_1}
    \end{subfigure}
    \hfill
    \begin{subfigure}[b]{0.49\linewidth}
         \centering
         \begin{subfigure}[b]{0.15\linewidth}
            \includegraphics[width=\textwidth,trim=300 20 150 30,clip]{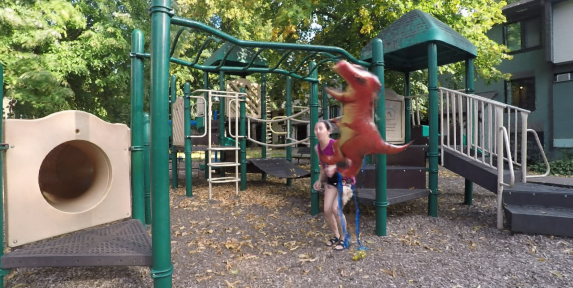}
            \caption*{RGB}
         \end{subfigure}
         \begin{subfigure}[b]{0.15\linewidth}
         \includegraphics[width=\textwidth,trim=300 20 150 30,clip]{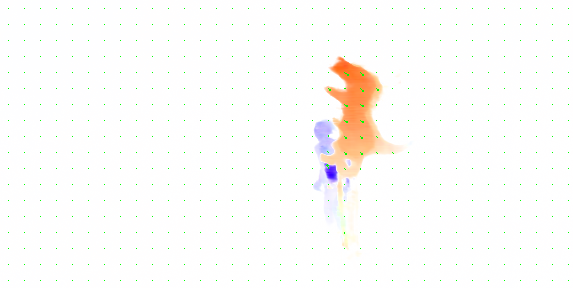}
         \caption*{$\mathbf{v}$}
         \end{subfigure}
         \begin{subfigure}[b]{0.15\linewidth}
         \includegraphics[width=\textwidth,trim=300 20 723 30,clip]{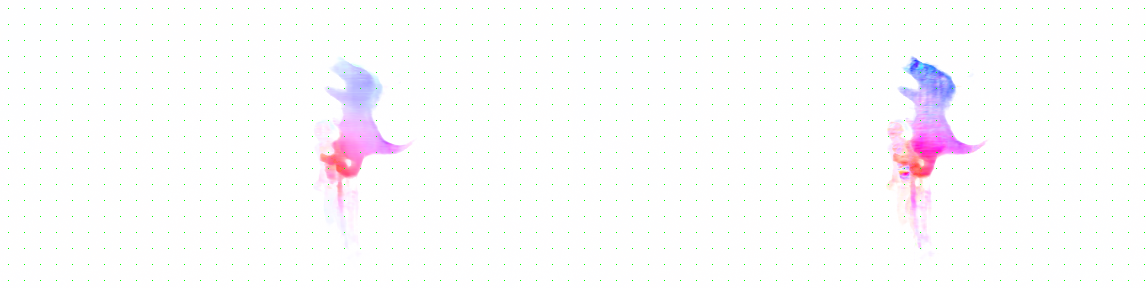}
         \caption*{$\mathbf{a}$}
         \end{subfigure}
         \begin{subfigure}[b]{0.15\linewidth}
         \includegraphics[width=\textwidth,trim=873 20 150 30,clip]{fig/kine_effect/069999_016_acc_fw-3.png}
         \caption*{Adv. $\mathbf{a}$}
         \end{subfigure}
         \begin{subfigure}[b]{0.15\linewidth}
         \includegraphics[width=\textwidth,trim=300 20 723 30,clip]{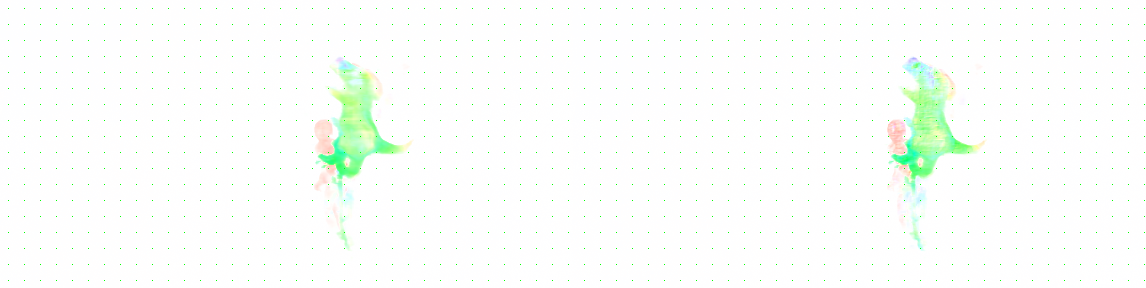}
         \caption*{$\mathbf{j}$}
         \end{subfigure}
         \begin{subfigure}[b]{0.15\linewidth}
         \includegraphics[width=\textwidth,trim=873 20 150 30,clip]{fig/kine_effect/test016_jer_fw.png}
         \caption*{Adv. $\mathbf{j}$}
         \end{subfigure}
         \caption{With kinematic regularization}
         \label{fig:kinematic_2}
    \end{subfigure}
    \caption{\textbf{Effect of kinematic regularization.} We visualize the rendered RGB and motion of each field. Without kinematic regularization, motion fields tend to show granular patterns. Our kinematic fields not only make the field smoother but also satisfy the kinematic property, \textit{i.e.}, $\mathbf{a} = \partial \mathbf{v}/\partial t + \mathbf{v}\cdot\nabla \mathbf{v}$. We abbreviate the advective equation to `Adv.' in the figure.
    }
    \label{fig:kinematic}
\end{figure}

\subsection{Kinematic Regularization}
Given the kinematic field, it is important to make sure the learned 
motion dynamics are physically plausible. As Fig.~\ref{fig:kinematic} indicates,
learning kinematic fields without adequate regularization may result in a sub-optimal kinematic field.
The subsequent sections elucidate regularization strategies.

\noindent\textbf{Integrity of kinematic fields.}
Learning the kinematic field $\mathcal{F}_\text{K}$ naively does not
guarantee the integrity of the kinematic quantities.
During training, the quantities are supervised via
the photometric consistency loss (Eq.~\ref{eq:photo_loss}),
which is applied upon the approximated displacement (Eq.~\ref{eq:taylor}).
It is important to note that even with supervision by the displacement, the physical relationships among the kinematic quantities may not be correct,
since the higher-order terms in the displacement equation (Eq.~\ref{eq:taylor}) make the system underdetermined. 

Given a velocity field, we can derive the acceleration field using the principle of advective acceleration. Given acceleration $\mathbf{a}$ and velocity $\mathbf{v}$ at a spatio-temporal coordinate,
the partial derivative and advection yield $\mathbf{a} = {\partial \mathbf{v}}/{\partial t} + \mathbf{v} \cdot \nabla \mathbf{v}$.
Similarly, we can formulate the relationship between $\mathbf{v}$, $\mathbf{a}$ and $\mathbf{j}$ as: $\mathbf{j} = {\partial \mathbf{a}}/{\partial t} + \mathbf{v} \cdot \nabla \mathbf{a}$.

To ensure kinematic consistency, 
we introduce the kinematic loss $\mathcal{L}_\text{K}$ defined as:
\begin{equation}
    \mathcal{L}_\text{K} = 
    \bigg\|\mathbf{a} - 
        \frac{\partial \mathbf{v}}{\partial t} - \mathbf{v} \cdot \nabla \mathbf{v}
    \bigg\|_2^2 + 
    \bigg\|\mathbf{j} - 
        \frac{\partial \mathbf{a}}{\partial t} - \mathbf{v} \cdot \nabla \mathbf{a}
    \bigg\|_2^2 + ...,
    \label{eq:kinematic_integrity}
\end{equation}
where $\nabla$ is the Jacobian operation.
To compute the partial derivatives and Jacobian matrices, we use a numerical method\footnote{$f'(x)= \frac{f(x+\epsilon) - f(x-\epsilon)}{2\epsilon}$} as presented in a neural surface reconstruction work~\cite{li2023neuralangelo}.

\noindent\textbf{Physics-informed regularization.}
In the context of regularizing radiance fields through the use of an auxiliary kinematic field, it is important to acknowledge the highly underdetermined nature of the problem. 
Essentially, there exists a broad spectrum of possible potential kinematic field solutions that could perfectly fit the observed training views. 
This complexity necessitates the introduction of a regularization strategy that can guide the motion representation toward more accurately reflecting real-world physics.

While not universally ideal,
rigidity~\cite{park2021nerfies} and
incompressiblility~\cite{chu2022physics,baieri2023fluid} have been identified as effective metrics for encapsulating real-world dynamics; 
in videos that we normally encounter, objects in a scene have nearly consistent density and are mostly rigid in a short time range.

\noindent \textbf{Rigidity.}
One benefit of having a velocity field is that we can measure the rate of deformation by its derivatives.
A strain rate tensor 
$D$ 
defines a kinematic property of a deformation;
it captures the rate at which a material undergoes deformation, and it is crucial for ensuring that objects maintain their structural integrity during motion.

In this paper, we propose
using the invariants of strain rate tensor~\cite{soria1994study,hart2010application} as a measure of rigidity.
The velocity gradient $\nabla \mathbf{v}$ sampled from the kinematic field, leads to the strain rate tensor $D$ by:
\begin{equation}
    D = \frac{1}{2} \left( \nabla \mathbf{v} + (\nabla \mathbf{v})^T \right).
\end{equation}
The rigidity loss function $\mathcal{L}_R$ is formulated based on the first and second invariants of the strain rate tensor:
\begin{equation}
    \mathcal{L}_R = \underset{\text{1st invariant}}{\underbrace{\lambda_\text{div}(\nabla \cdot\mathbf{v})^2}} + 
    \underset{\text{2nd invariant}}{\underbrace{\big(\frac{1}{2} (\text{tr}(D)^2 - \text{tr}(D\cdot D))\big)^2}} , \label{eq:strain}
\end{equation}
where $\text{tr}(\cdot)$ denotes the trace operator and $\lambda_\text{div}$ controls the balance between the two terms.
Note that
the first invariant of strain rate measures dilational change (\textit{i.e.}, volumetric change)  with respect to time and
the second invariant of strain rate measures distortional change.

\begin{figure}[t]
    \centering
    \begin{minipage}{0.432\textwidth}
        \centering
        \includegraphics[trim={0.5cm 0.5cm 0cm 0cm},clip,width=\linewidth]{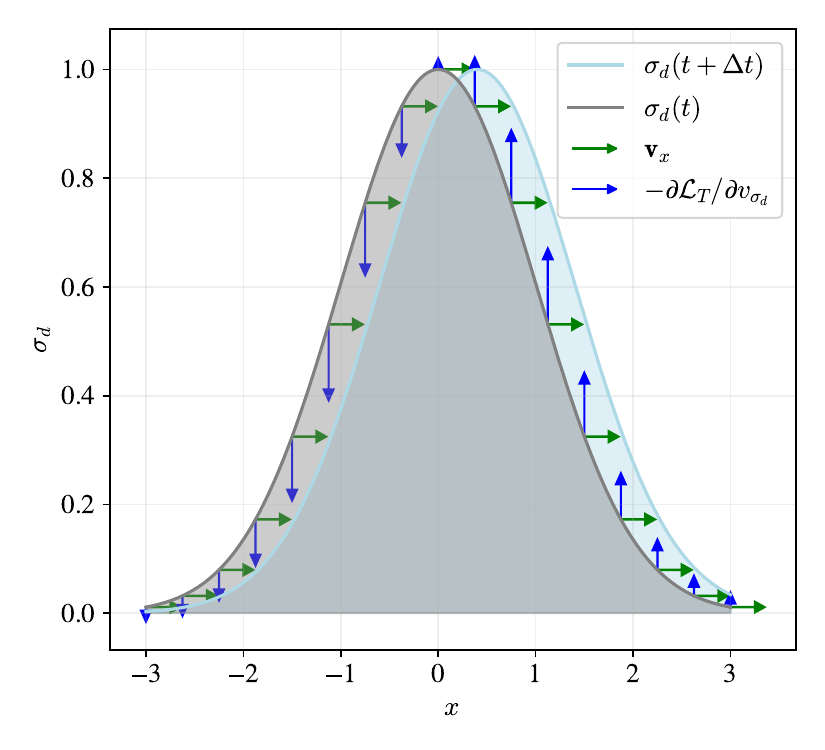}
        \caption{\textbf{Visualization of density variation by a spatial coordinate $x$.} The plot displays the gradient $-\partial \mathcal{L}_\text{T}/\partial v_{\sigma_d}$ with arrows.
        With the velocity field directed to the right (\textit{i.e.}, $\mathbf{v}_x = 0.3$), we can compute the gradient of the transport regularization $\mathcal{L}_\text{T}$ w.r.t. the rate of density change $v_{\sigma_d} = \partial \sigma_d / \partial t$.
        Minimizing $\mathcal{L}_\text{T}$ allows us to render the density at $t+\Delta t$ aligned with the flow field.}
        \label{fig:transport_graph}
    \end{minipage}\hfill
    \begin{minipage}{0.540\textwidth}
        \centering
        \includegraphics[trim={13.6cm 42.2cm 64.8cm 16cm},clip,width=\linewidth]{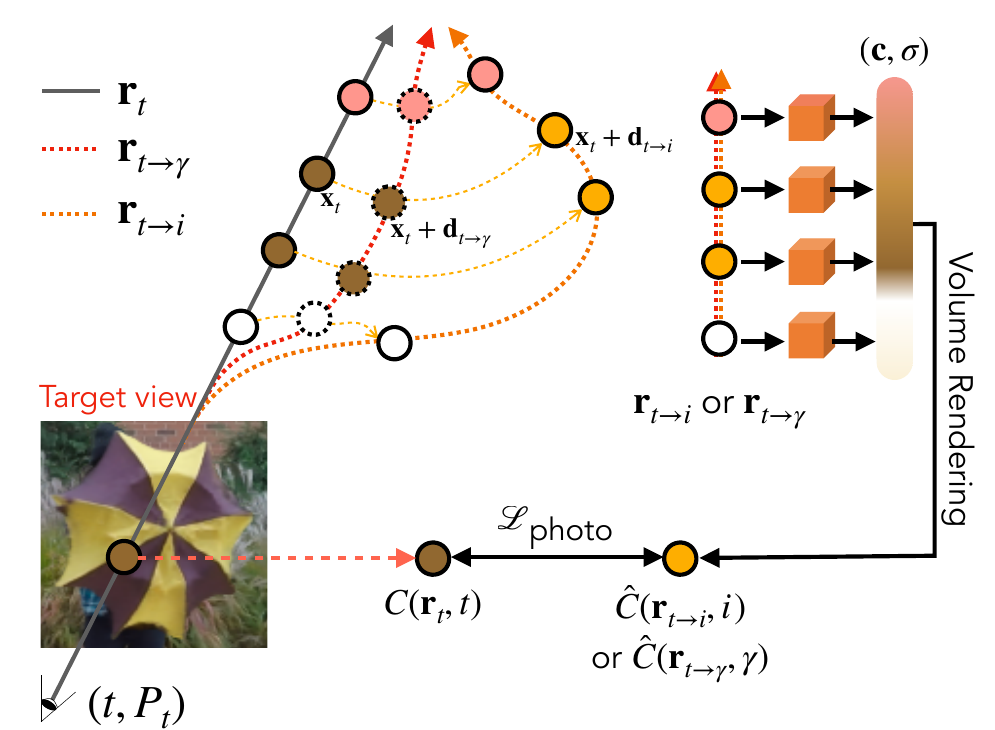} 
        \caption{\textbf{Photometric consistency loss.}
        During optimization, we deform each ray from a reference time and pose $(t, P_t)$ to different timestamps $\gamma$ and $i$.
        Here, we sample $i$ from the neighboring frame times, and we consequently sample an intermediate timestamp $\gamma\sim \mathcal{U}(t, i)$. Given each timestamp, we can deform a ray using Eq.~\ref{eq:taylor}.
        The photometric loss $\mathcal{L}_\text{photo}$
          is computed based on the color consistency between the deformed ray and the original color, enhancing temporal consistency.}
        \label{fig:photo}
    \end{minipage}
\end{figure}

\noindent \textbf{Transport.}
A transport equation has been adopted in physics-informed neural networks~\cite{chu2022physics}.
By simplifying the Navier-Stokes equation, a transport loss can be defined as:
$\frac{\partial \sigma}{\partial t} + \mathbf{v} \cdot \nabla \sigma=0$.
We leverage the physics-informed regularizers based on the transport equation and the divergence of the velocity:
\begin{align}
    \mathcal{L}_\text{T} &= 
    \bigg(
        \frac{\partial \sigma_d}{\partial t} + \mathbf{v} \cdot \nabla \sigma_d
    \bigg)^2, \label{eq:transport} 
\end{align}
where $\sigma_d$ is the volume density used for volume rendering.
This aligns the velocity field with the density field (\textit{i.e.}, Fig.~\ref{fig:transport_graph}), preventing a sudden disappearance or appearance of objects 
in the dynamic radiance field.

\subsection{Learning Objectives}

\noindent \textbf{Cycle consistency.}
The cycle consistency is essential to ensure that the trajectories derived from the kinematic field do not introduce misalignments~\cite{li2021neural}.
We use cycle constraints among four timestamps $t$, $i$, $j$, and $\gamma$.
Given a reference time $t$, we randomly sample time $i$ from the neighboring frames within a threshold. 
We define $j=t + 2(i - t)$, making it twice the distance from 
$t$ as 
$i$ is from $t$.
$\gamma\sim\mathcal{U}(t, i)$ is an intermediate time, uniformly sampled between $t$ and $i$.
With the time variables, we formulate our loss:
\begin{align}
\begin{aligned}
    \mathcal{L}_\text{C} = \rho(\mathbf{d}_{t\rightarrow i} + \mathbf{d}_{i\rightarrow t})
    +\rho(\mathbf{d}_{t\rightarrow i \rightarrow j} - \mathbf{d}_{t\rightarrow j})
    +\rho(\mathbf{d}_{t\rightarrow \gamma \rightarrow i} - \mathbf{d}_{t\rightarrow i}),\label{eq:cycle_loss}
\end{aligned}
\end{align}
where $\rho$ denotes the generalized chabonnier loss~\cite{sun2014quantitative}. Here, we abbreviate the displacement $d(\cdot)$ (Eq.~\ref{eq:taylor}) for conciseness: $\mathbf{d}_{t\rightarrow i}$ is shorthand for $d(\mathbf{x},t,i-t)$, and $\mathbf{d}_{t\rightarrow i \rightarrow j}$ simplifies $d(\mathbf{x} + \mathbf{d}_{t\rightarrow i}, i, j -i)$.

\noindent \textbf{Photometric consistency loss.}
To enhance temporal coherency, we deform rays into nearby frame times and enhance the consistency among them (Fig.~\ref{fig:photo}). Specifically, given a reference ray $\mathbf{r}_t$ at time $t$, a pixel color $\hat{C}(\mathbf{r}_t, t)$ can be obtained using the volume rendering (Eq.~\ref{eq:vol_render}). 
In addition to the reference ray, we apply ray deformation based on the predicted kinematic quantities. As in cycle consistency (Eq.~\ref{eq:cycle_loss}), 
we sample time $i$ from neighboring frames, and time $\gamma \sim \mathcal{U}(t, i)$ between times $t$ and $i$. 
Different from existing approaches~\cite{li2021neural,Gao_2021_ICCV},
we deform rays with kinematic quantities using Taylor approximation.
The deforming process results in two deformed rays $\mathbf{r}_{t\rightarrow i}$ and $\mathbf{r}_{t\rightarrow \gamma}$, and two rendered colors $\hat{C}(\mathbf{r}_{t\rightarrow i}, i)$ and $\hat{C}(\mathbf{r}_{t\rightarrow \gamma}, \gamma)$, correspondingly.
Based on the deformed rays, we apply the photometric consistency loss, comparing the rendered color with the ground-truth color.
The loss function is formulated as:
\begin{align}
\begin{aligned}
    \mathcal{L}_\text{photo} &= 
    \|\hat{C}(\mathbf{r}_t, t)-C(\mathbf{r}_t, t)\|_2^2\\
    &+\lambda \alpha(\mathbf{r}_t) \rho(\hat{C}(\mathbf{r}_{t\rightarrow i}, i)-C(\mathbf{r}_t, t))
    +\lambda\alpha(\mathbf{r}_t)\rho(\hat{C}(\mathbf{r}_{t\rightarrow \gamma}, \gamma)-C(\mathbf{r}_t, t)),\label{eq:photo_loss}
\end{aligned}
\end{align}
where $\lambda$ is a loss weight for the temporal consistency, and $\alpha(\mathbf{r}_t)$ is the sum of rendering weights of the dynamic densities, representing the dynamic likelihood of the ray $\mathbf{r}_t$.

\noindent \textbf{Final objective.}
Our final objective function, denoted as $\mathcal{L}$, is the weighted sum of the various loss terms discussed previously.
To keep things simple with the variety of loss functions, we symbolize the losses to $\mathcal{L}_\text{photo}$, $\mathcal{L}_\text{kinematic}$, and $\mathcal{L}_\text{reg}$. 
This results in the final loss:
\begin{equation}
    \mathcal{L} = \mathbb{E}_{\mathbf{r}_t}\big[\mathcal{L}_\text{photo} + \mathcal{L}_\text{kinematic} + \mathcal{L}_\text{reg}\big].\label{eq:final_loss}
\end{equation}
Here, $\mathcal{L}_\text{kinematic}$ is the weighted sum of the kinematic integrity $\mathcal{L}_\text{K}$, transport $\mathcal{L}_\text{T}$, and rigidity $\mathcal{L}_\text{R}$.
Note that $\mathbb{E}_{\mathbf{r}_t}[\mathcal{L}_\text{kinematic}]$ is computed with randomly sampled points along the ray $\mathbf{r}_t$ from each minibatch.
The additional regularization $\mathcal{L}_\text{reg}$ includes the skewed entropy $\mathcal{L}_b$, the cycle consistency $\mathcal{L}_\text{C}$, 
total-variation loss~\cite{chen2022tensorf,cao2023hexplane}, 
distortion loss~\cite{barron2021mip}, 
monocular depth supervision~\cite{li2021neural}, 
normal regularization~\cite{Jin2023TensoIR,verbin2022ref}
and optical flow supervision~\cite{li2021neural}.

%% file: sec/4_experiment.tex
\section{Experiment}

\subsubsection{Dataset.} 
We use the NVIDIA dynamic view synthesis dataset (NDVS)~\cite{yoon2020novel} to evaluate the effectiveness of our model.
For the monocular setup, we adopt the preprocessing steps proposed by NSFF~\cite{li2021neural} and DynamicNeRF~\cite{Gao_2021_ICCV}.
This dataset has 8 scenes and is captured with 12 synchronized cameras.
When used for a monocular video reconstruction task, 
the dataset shows relatively low effective multi-view factors (EMFs)~\cite{gao2022monocular}, compared to other monocular datasets;
this means that the dataset is more challenging to reconstruct due to the less camera motion and faster-moving objects.

There are two widely used monocular NDVS protocols: \textbf{12-frames}~\cite{Gao_2021_ICCV} and \textbf{24-frames}~\cite{li2021neural}.
Most papers opt for either protocol; however, we include both protocols for rigorous evaluation.
Meanwhile, \textbf{24-frames-sparse} is proposed by us, as neither existing protocol provides novel time evaluation; we leave out odd-numbered frames at unseen times for testing. 
For the `Dynamic Only' marks in tables, we use the dynamic mask to evaluate only in terms of dynamic pixels; please refer to \cite{li2021neural,Gao_2021_ICCV} for how to generate dynamic masks. We detail the protocols below.

\begin{itemize}
    \item \noindent\textbf{24-frames}: this sequence is configured with 24 training frames, and 264 test frames from unseen multi-views.
    For a fair comparison, we utilize the dynamic mask, optical flow, and monocular depth, extracted by the same model with the prior work~\cite{li2021neural}.

    \item \noindent\textbf{24-frames-sparse}:
    we reduce the 24-frames sequence into 12 frames by sampling the even-numbered frames for training. This sparse sampling allows for testing novel view synthesis at both seen times and novel times.

    \item \noindent\textbf{12-frames}:
    this sequence has 12 training and 12 testing samples.
    We use the preprocessed dynamic mask, optical flow, and monocular depth downloadable from the code repository of DynamicNeRF~\cite{Gao_2021_ICCV}.
\end{itemize}

\subsection{Implementation Details}
\noindent\textbf{Fields configuration.} 
Our radiance and kinematic fields are configured with HexPlane~\cite{cao2023hexplane} and TensoRF~\cite{chen2022tensorf}.
After features are sampled from the point queries,
small MLP decoders process the features to yield colors and kinematic outputs. For density outputs, we use a linear layer following the practices in \cite{cao2023hexplane,chen2022tensorf}.
Further details on field configurations are available in the supplementary material.

\noindent\textbf{Training.}
We use 70,000 steps to optimize our models, which takes about 7~hours with one NVIDIA V100 or A100 GPU. 
We initialize the static field for 5,000~steps, taking approximately 5~minutes. After finishing the initialization of the static field, we jointly train the fields: dynamic, static, and kinematic fields. During the joint training, we lower the learning rate of the static field by the factor of 0.1 at predefined iterations
to prevent dynamic objects from being portrayed as static objects.
Additional training details are provided in the supplementary material.

\noindent \textbf{Evaluation metrics.}
We report peak signal-to-noise ratio (PSNR), structural similarity (SSIM), and perceptual similarity (LPIPS).
These three metrics have been used as standard in prior studies to evaluate the quality of renderings. 

\noindent \textbf{Base architecture (HexPlane).}
Since our model is based on tensor-decomposed radiance fields, we compare our models with HexPlane~\cite{cao2023hexplane}.
The baseline model is the dynamic radiance field $\mathcal{F}_\text{DY}$, without the static radiance field $\mathcal{F}_\text{ST}$ and the kinematic field $\mathcal{F}_\text{K}$.
For a fair comparison, we use monocular depth supervision both to the baseline model and our models, although the original HexPlane model does not support 
the supervision from monocular depth. 
Since the base HexPlane model does not have an auxiliary flow field, we do not use the optical flow supervision and the photometric consistency loss.

\input{tab/ablation}
\begin{figure}[t]
    \centering
    \includegraphics[trim={0cm 34.5cm 36.37cm 0cm},clip,width=0.95\linewidth]{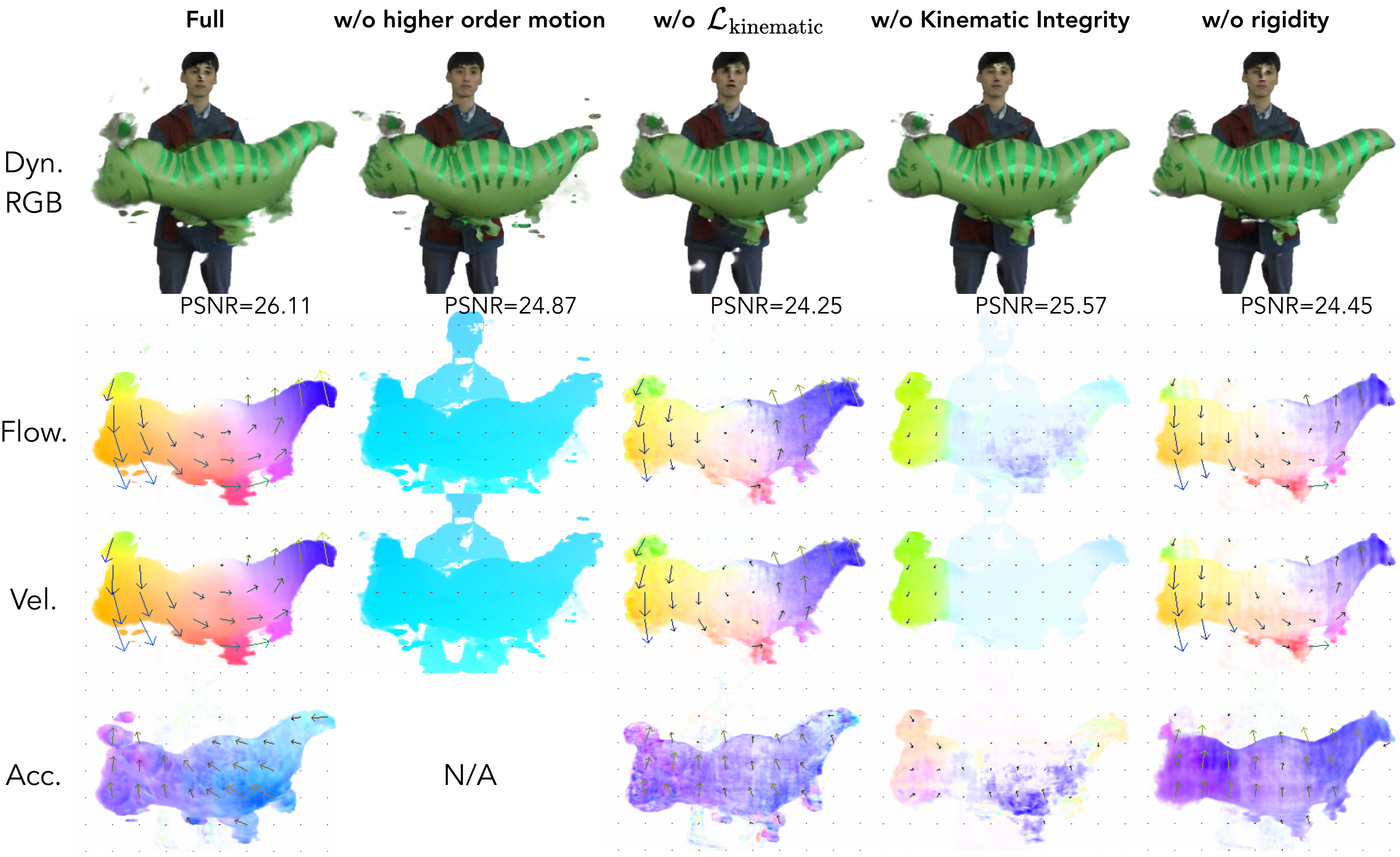}
    \caption{\textbf{Comparative visualization of kinematic fields.}
    The first row shows RGB values rendered from our dynamic radiance field, showing PSNR values from a ground truth RGB image.
    From the second to the last row, we show flow (\textit{i.e.}, displacement), velocity, and acceleration.
    }
    \label{fig:ablation}
\end{figure}

\subsection{Evaluation of Kinematic Fields}
\noindent\textbf{Ablation of kinematic losses.}
To show the effectiveness of our approach, we conduct a comprehensive ablation study on the Balloon1 scene of the NDVS dataset, focusing on kinematic loss components.
In Table~\ref{tab:ablation12}, we show ablative results from seen and unseen frame times.
Our findings indicate that the model achieves optimal performance when all components are integrated.

Fig.~\ref{fig:ablation} delves into the kinematic field analysis. 
Omitting higher order motion restricts the model to utilizing only the velocity field, which proves insufficient for capturing non-linear motion. This limitation hinders the model's ability to minimize kinematic losses, leading to a suboptimal flow representation.
In the absence of $\mathcal{L}_\text{kinematic}$,
the radiance field suffers from reduced PSNR scores, which can be attributed to the physically incorrect motion and geometry.
Compared to the full model, the model without the rigidity loss results in noisy, non-rigid motion, resulting in incorrect motion and the worse reconstruction accuracy.

\input{fig/ord/ord}

\noindent\textbf{Motion order.}
Our study represents motion through various kinematic quantities.
Adjusting the kinematic field to include different orders of motion, we can model our system using just velocity~\cite{du2021neural,chu2022physics}, or extend to higher orders -- fourth (snap) and fifth (crackle) -- by extending the Taylor series (Eq.~\ref{eq:taylor}) and the kinematic integrity (Eq.~\ref{eq:kinematic_integrity}) equations. 
Fig.~\ref{fig:motion_order} illustrates the influence of motion order on reconstruction accuracy for the NDVS dataset. 
As motion order increases, the reconstruction accuracy first improves, peaking at jerk, 
but further orders degrades the accuracy.
We conjecture that the augmentation of model complexity with higher orders of kinematic quantities may render the model more susceptible to overfitting.

\input{tab/comparison_average}

\subsection{Comparison with SOTA models}

\textbf{Quantitative results.}
NSFF~\cite{li2021neural} is a dynamic neural radiance field model, which utilizes scene-flow fields for the regularization of a dynamic radiance field.
Since NSFF is a neural model, the network training takes about 2 days on two V100 GPUs (96 GPU hours), using the official code by the authors.
Note that optimizing a scene using our method takes about 7 hours per scene, which makes ours 12 times faster than NSFF in terms of GPU hours.
We compare our results with NSFF~\cite{li2021neural} in Table~\ref{tab:comparison_24frames_avg}.
In the table, our method shows better accuracy in terms of all metrics.
In addition, we report results on NDVS (12-frames) in Table~\ref{tab:comparison_12frames_avg}.

\begin{figure}[t]
    \centering
    \includegraphics[trim={0cm 77.8cm 47.05cm 0cm},clip,width=1.0\linewidth]{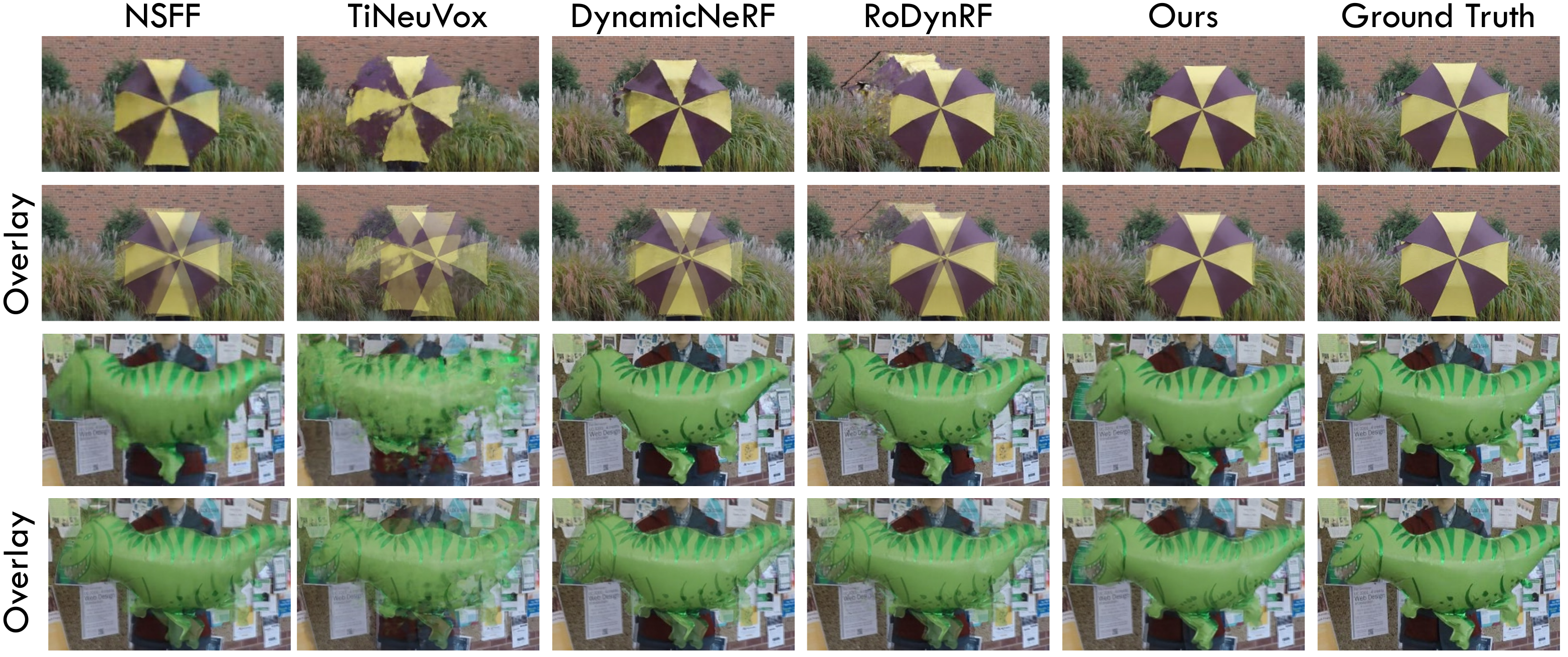}
    \caption{\textbf{Qualitative results} on NDVS (12-frames).
    The second and fourth rows illustrate the overlay of the synthesized and ground truth views of the scene.
    }
    \label{fig:qual_sota}
\end{figure}

\noindent\textbf{Qualitative results.} 
Fig.~\ref{fig:qual_sota} illustrates qualitative results, including the synthesized images, ground truth, and their overlaid images.
As shown in the first and third rows, our method improves the visual quality of the rendered views.
Additionally, the overlaid images demonstrate the geometrical accuracy of our model, showcasing its better alignment with the ground truth. 
This enhanced performance is attributed to the kinematics and physics incorporated in our methodology, which is fundamental to the improvement.

%% file: tab/ablation.tex
\begin{table}[t]
\caption{\textbf{Ablation study} on Balloon1 scene of the NDVS dataset (24-frames-sparse). Please refer to the supplementary material for ablation results on other NDVS scenes.}
\resizebox{\linewidth}{!}{
\begin{tabular}{l|cccccc|cccccc}
\hline
&\multicolumn{6}{c|}{Novel Times} & \multicolumn{6}{c}{Seen Times}   \\
&\multicolumn{3}{c}{Full} & \multicolumn{3}{c|}{Dynamic Only} &\multicolumn{3}{c}{Full} & \multicolumn{3}{c}{Dynamic Only}  \\
Methods & PSNR $\uparrow$ & SSIM $\uparrow$ & LPIPS $\downarrow$ & PSNR $\uparrow$ & SSIM $\uparrow$ & LPIPS $\downarrow$ & PSNR $\uparrow$ & SSIM $\uparrow$ & LPIPS $\downarrow$ & PSNR $\uparrow$ & SSIM $\uparrow$ & LPIPS $\downarrow$\\\hline
HexPlane~\cite{cao2023hexplane} & 
19.84&	0.668&	0.145&	18.12&	0.419&	0.251& 18.93&	0.649&	0.137&	17.74&	0.429&	0.202  \\\hline
\textbf{Ours} & 
\textbf{25.67}&	\textbf{0.865}&	\textbf{0.064}&	\textbf{21.16}&	\textbf{0.625}&	\textbf{0.170}& \textbf{25.80}&	\textbf{0.867}&	\textbf{0.050}&	\textbf{21.35}&	\textbf{0.632}&	\textbf{0.121}\\\hline
w/o Higher order motion &
24.36&	0.842&	0.077&	19.93&	0.556&	0.200& 24.63&	0.846&	0.061&	20.22&	0.572&	0.145 \\
w/o $\mathcal{L}_\text{kinematic}$ & 
24.96&	0.853&	0.068&	20.15&	0.568&	0.185 & 25.30&	0.858&	0.052&	20.65&	0.594&	0.129\\
w/o Kinematic integrity & 
25.46&	0.862&	0.066&	20.78&	0.607&	0.179 & 25.57&	0.865&	0.051&	21.02&	0.620&	0.126\\
w/o Rigidity &  
25.11&	0.856&	0.066&	20.40&	0.582&	0.181 & 25.43&	0.860&	0.051&	20.82&	0.603&	0.123 \\\hline

\end{tabular}
}

\label{tab:ablation12}
\end{table}

%% file: fig/ord/ord.tex
\begin{figure}[t]
    \centering
    \begin{subfigure}[b]{0.24\linewidth}
        \includegraphics[width=\linewidth]{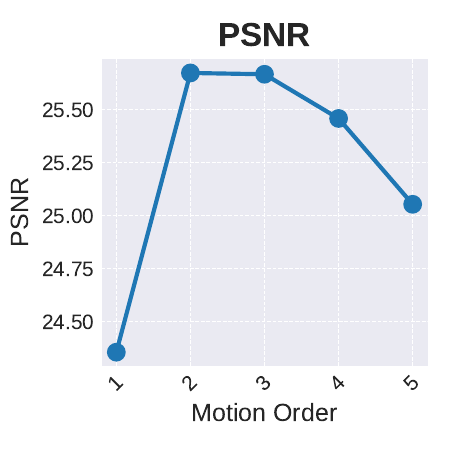}
    \end{subfigure}
    \begin{subfigure}[b]{0.24\linewidth}
        \includegraphics[width=\linewidth]{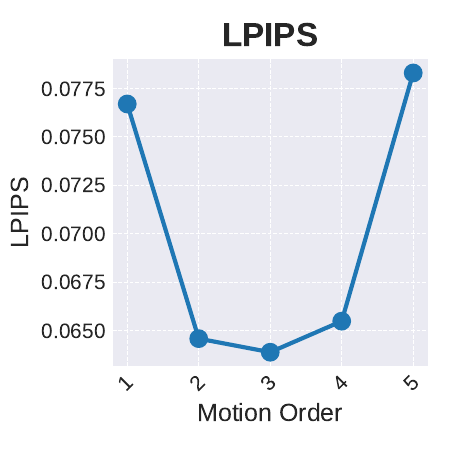}
    \end{subfigure}
    \begin{subfigure}[b]{0.24\linewidth}
        \includegraphics[width=\linewidth]{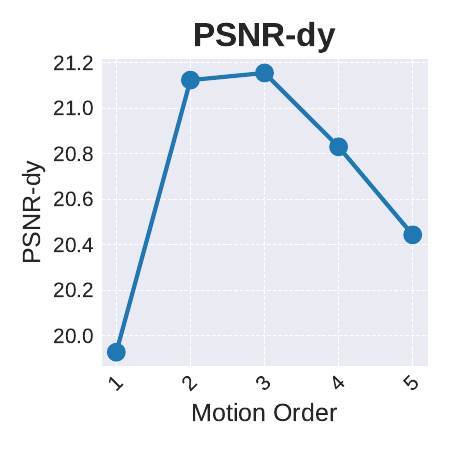}
    \end{subfigure}
    \begin{subfigure}[b]{0.24\linewidth}
        \includegraphics[width=\linewidth]{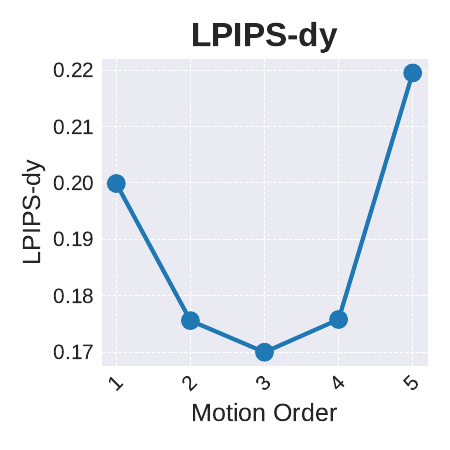}
    \end{subfigure}
    \caption{\textbf{Motion Order}. Reconstruction Accuracy for the Balloon1 Scene (24-frames-sparse): The $x$-axis of each plot categorizes maximum motion order as follows: velocity (1), acceleration (2), jerk (3), snap (4), and crackle (5). The metric utilized for each plot is specified at the top.}
    \label{fig:motion_order}
\end{figure}

%% file: tab/comparison_average.tex
\begin{table}[t]
    \centering
    \caption{\textbf{Quantitative comparison} on NDVS. This table reports average results on all scenes. Please refer to the supplementary material for per-scene results.}
    \begin{subtable}[h]{0.363\textwidth}
        \centering
        \caption{12-frames}\label{tab:comparison_12frames_avg}
        \resizebox{\linewidth}{!}{
        \begin{tabular}{lcc}
        \hline 
        & \multicolumn{2}{c}{Full}\\
        Methods&PSNR$\uparrow$&LPIPS$\downarrow$ \\\hline
        NeRF~\cite{mildenhall2020nerf}* & 21.99&0.250 \\
        NSFF~\cite{li2021neural}* &24.33&0.199 \\
        DynamicNeRF~\cite{Gao_2021_ICCV}* &\underline{26.10}&0.082 \\
        HyperNeRF~\cite{park2021hypernerf}* &17.60&0.367 \\
        RoDynRF~\cite{Liu_2023_CVPR}* & 25.89&\textbf{0.065} \\\hline
        Ours & 
        \textbf{26.33}&\underline{0.071}\\\hline
        \multicolumn{3}{l}{*Results reported in \cite{Liu_2023_CVPR}}
        \end{tabular}}
    \end{subtable}\hfill
    \begin{subtable}[h]{0.629\textwidth}
        \centering
        \caption{24-frames}\label{tab:comparison_24frames_avg}
        \resizebox{\linewidth}{!}{
        \begin{tabular}{lcccccc}
        \hline
         & \multicolumn{3}{c}{Full} & \multicolumn{3}{c}{Dynamic Only} \\
        Methods           & \multicolumn{1}{l}{PSNR$\uparrow$}      & \multicolumn{1}{l}{SSIM$\uparrow$}      & \multicolumn{1}{l}{LPIPS$\downarrow$}     & \multicolumn{1}{l}{PSNR$\uparrow$}      & \multicolumn{1}{l}{SSIM$\uparrow$} & \multicolumn{1}{l}{LPIPS$\downarrow$}     \\\hline
        NeRF~\cite{mildenhall2020nerf}*        &	24.90&	0.893&	0.098&	16.98&	0.532&	0.314\\
        Luo et al.~\cite{luo2020consistent}* &	21.37&	0.746&	0.141&	16.97&	0.530&	0.207\\
        Yoon et al.~\cite{yoon2020novel}*  &	21.78&	0.761&	0.127&	17.34&	0.547&	0.200\\
        NSFF~\cite{li2021neural}*     &	28.19&	0.928&	0.045&	21.91&	0.758&	0.097\\
        NSFF~\cite{li2021neural}      &	28.04&	0.927&	0.045&	21.34&	0.740&	0.111\\\hline
        Ours      &	\textbf{28.78}&	\textbf{0.936}&	\textbf{0.042}&	\textbf{22.13}&	\textbf{0.774}&	\textbf{0.094}\\\hline
        \multicolumn{7}{l}{*Results reported in \cite{li2021neural}}\\
        \end{tabular}
        }
        
    \end{subtable}
\end{table}

%% file: sec/5_conclusion.tex
\section{Conclusion}

In this paper, we have presented a novel method to learn dynamic radiance fields with kinematic fields.
We propose the kinematic field, where kinematic quantities are learned by joint learning with the radiance fields.
Using the kinematic quantities from the kinematic fields, we propose physics-based loss functions that underpin our method.
Lastly, we quantitatively and qualitatively show the effectiveness of our method on the monocular dataset.
We believe that this work not only introduces the potential of incorporating kinematics into the learning of radiance fields but also sets a foundation for future direction in physics-based dynamic scene understanding.

%% file: sec/6_supplementary.tex
\appendix
\newcommand{\hbAppendixPrefix}{A}
\renewcommand{\thesection}{\hbAppendixPrefix\arabic{section}}
\setcounter{section}{0}
\renewcommand{\thefigure}{\hbAppendixPrefix\arabic{figure}}
\setcounter{figure}{0}
\renewcommand{\thetable}{\hbAppendixPrefix\arabic{table}} 
\setcounter{table}{0}
\renewcommand{\theequation}{\hbAppendixPrefix\arabic{equation}} 
\setcounter{equation}{0}  

In this supplementary material, we provide additional experimental results. 
In addition, we describe details of our method 
including coordinate systems, and the tensor-decomposition-based architectures, and detailed training strategies.
Lastly, we discuss limitation and potential negative impact of our work.

\section{Additional Results}
In this section, we show additional results, which can supplement the main paper's results.

\subsection{Ablation on Different Scenes}
\noindent
\textbf{Quantitative results.}
In Table~\ref{tab:supp_ablation},
we report additional ablation study results on the NDVS (24-frames-sparse) scenes.
In most scenes, our full model consistently surpasses alternative approaches.
Especially, the effectiveness of our method is most evident within dynamic regions (\textit{i.e.}, Dynamic Only metrics) of the scenes, showing 
meaningful gaps over the ablated models. 

\begin{figure}[tp]
    \centering
    \includegraphics[page=1,width=\linewidth,trim=0 10.3cm 124.59cm 0,clip]{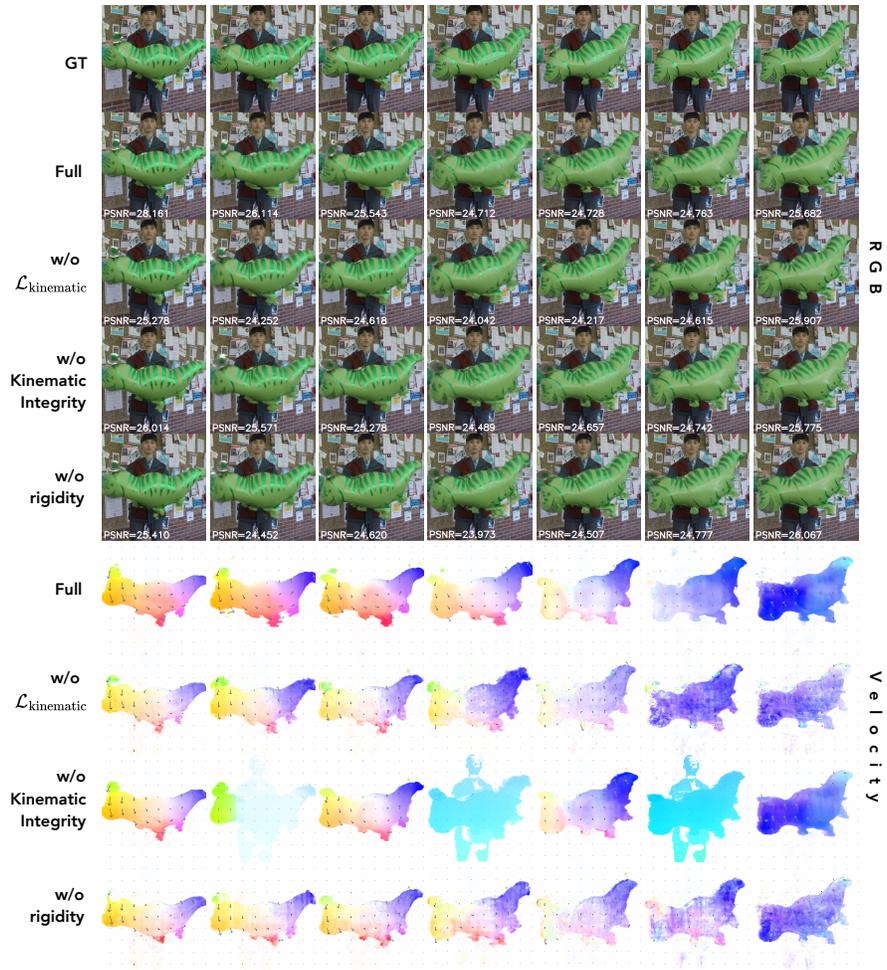}
    \caption{\textbf{Qualitative comparison} on Balloon1 scene of the NDVS dataset (24-frames-sparse)}
    \label{fig:supp_qual_balloon1}
\end{figure}

\begin{figure}[tp]
    \centering
    \includegraphics[page=4,width=\linewidth,trim=0 10.3cm 121.59cm 0,clip]{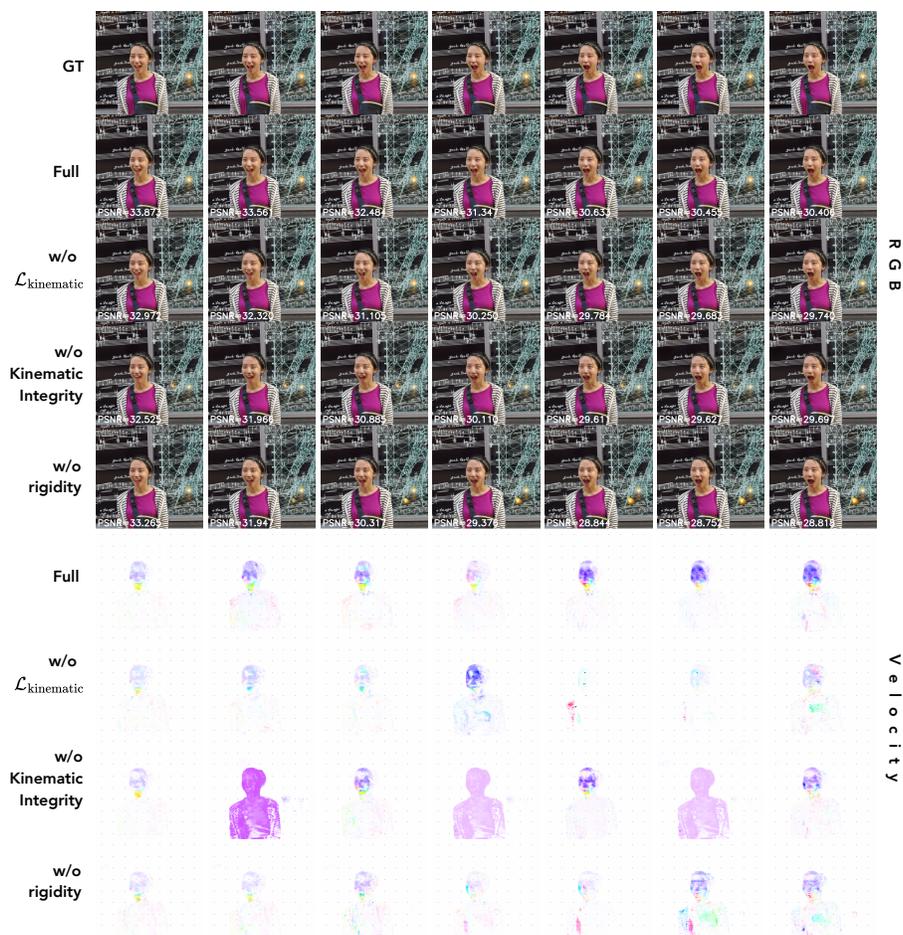}
    \caption{\textbf{Qualitative comparison} on DynamicFace scene of the NDVS dataset (24-frames-sparse)}
    \label{fig:supp_qual_dynface}
\end{figure}

\begin{figure}[tp]
    \centering
    \includegraphics[page=2,width=\linewidth,trim=0 10.3cm 124.59cm 0,clip]{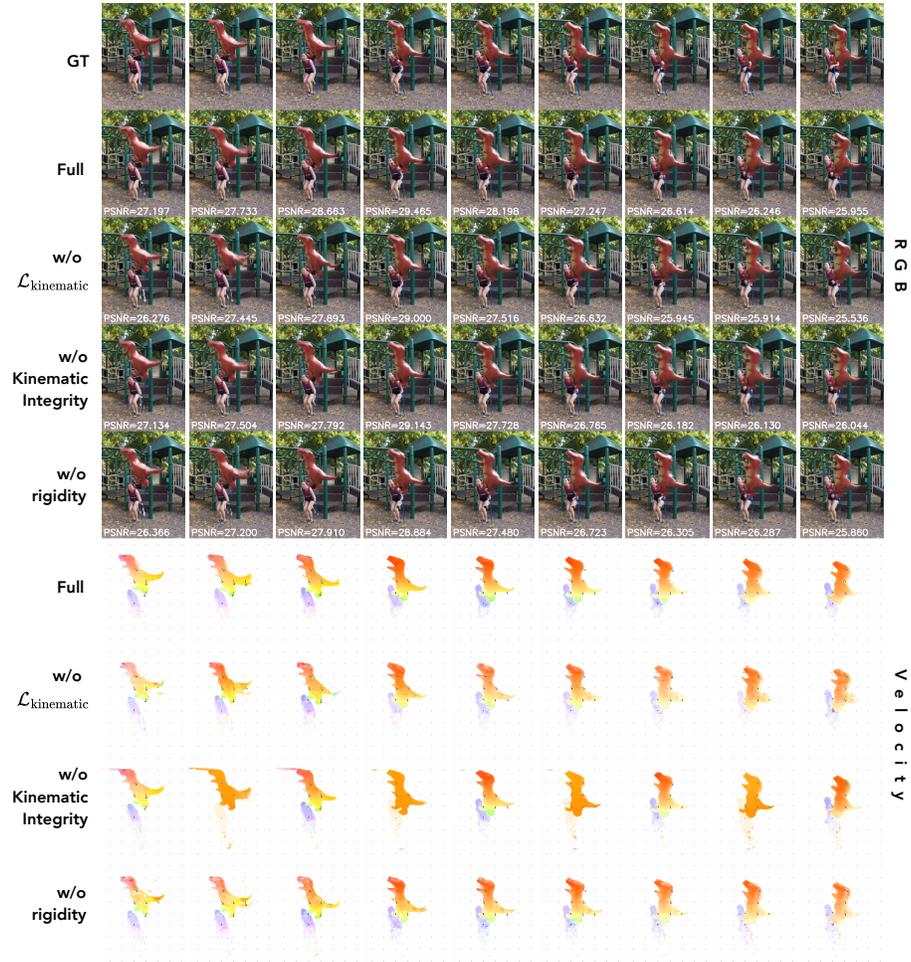}
    \caption{\textbf{Qualitative comparison} on Playground scene of the NDVS dataset (24-frames-sparse)}
    \label{fig:supp_qual_playground}
\end{figure}

\begin{figure}[tp]
    \centering
    \includegraphics[page=3,width=\linewidth,trim=0 32.89cm 124.59cm 0,clip]{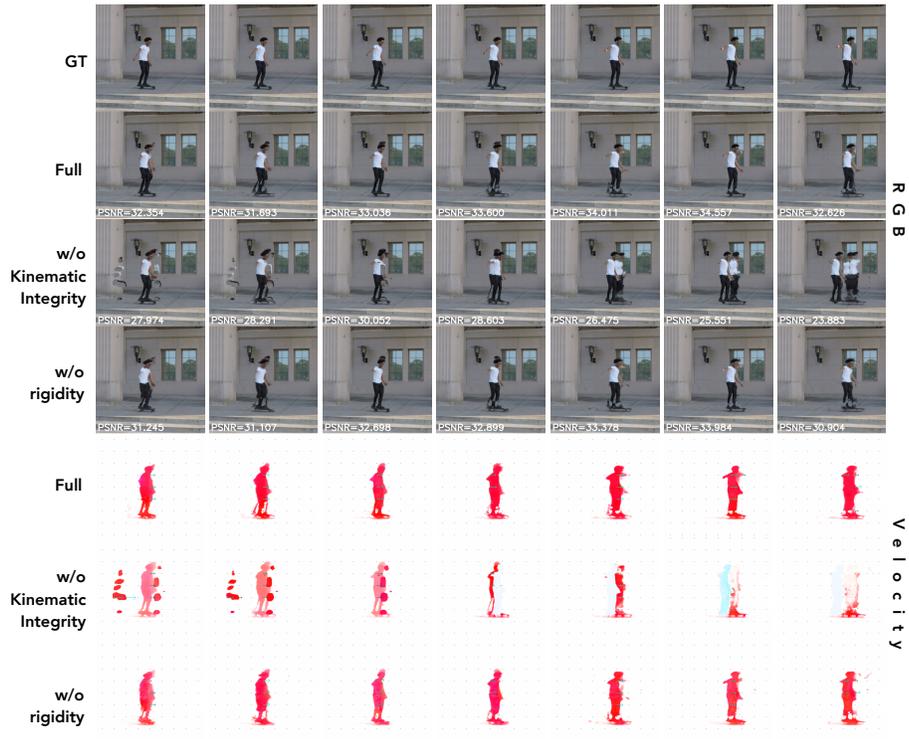}
    \caption{\textbf{Qualitative comparison} on Skating scene of the NDVS dataset (24-frames-sparse). Note that the w/o $\mathcal{L}_\text{kinematic}$ model did not converge in this scene.}
    \label{fig:supp_qual_skating}
\end{figure}

\begin{figure}[tp]
    \centering
    \includegraphics[width=0.95\linewidth,trim=0 8.85cm 40.4cm 0,clip]{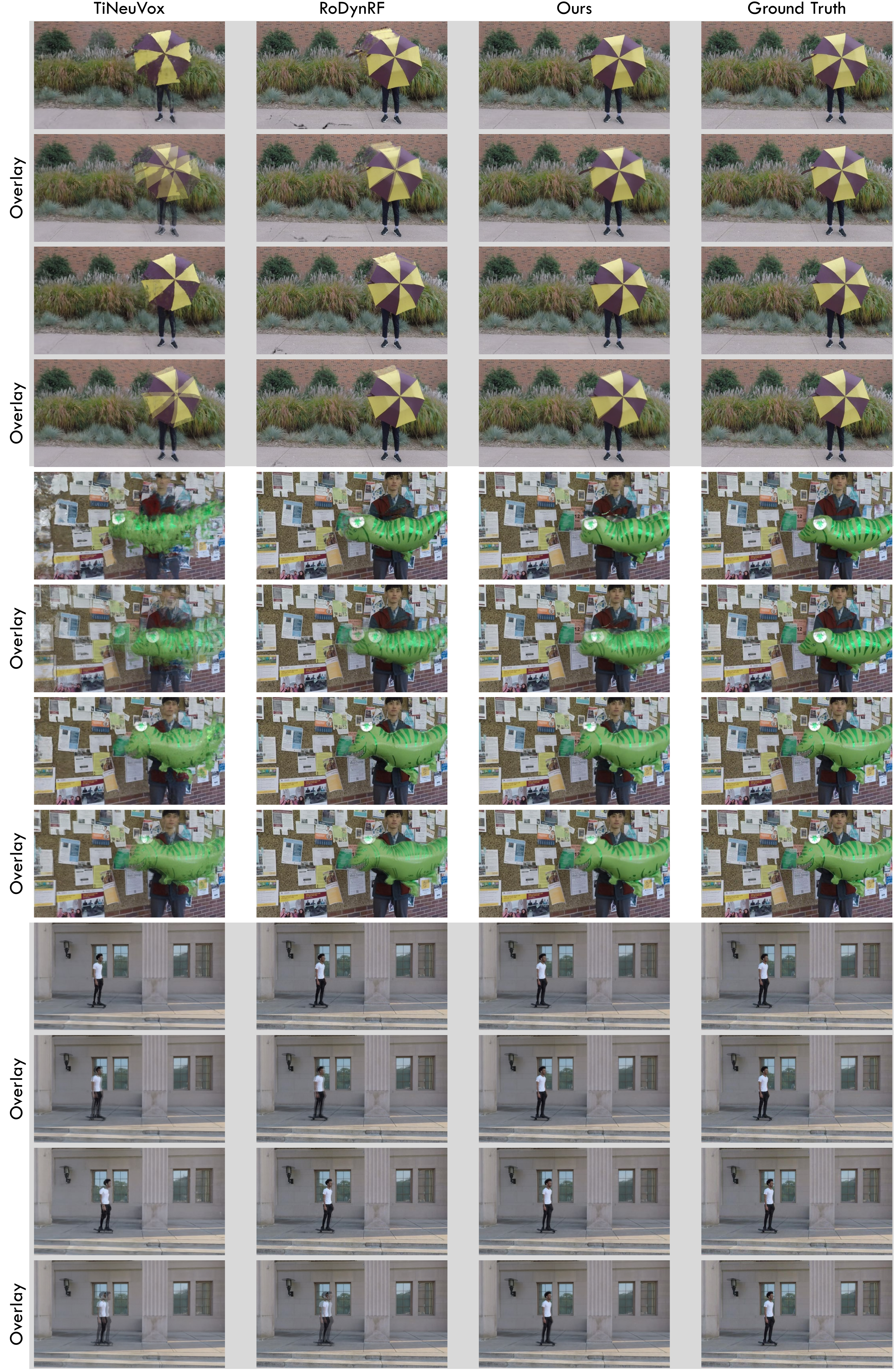}
    \caption{\textbf{Qualitative comparison} on NDVS dataset (12-frames) }\label{fig:qual_supp_12frames}
\end{figure}

\noindent
\textbf{Qualitative results.}
In Fig.~\ref{fig:supp_qual_balloon1}-\ref{fig:supp_qual_skating}, we visualize the rendered RGB maps and 
velocity fields inferred from ablated models.
In the figure, our full model not only shows better visual quality, but also presents the most smooth and consistent velocity field.
Notably, learning kinematic fields without the kinematic integrity loss results in highly inconsistent flow field.
In Fig.~\ref{fig:qual_supp_12frames}, we qualitatively compare ours with existing works.

\subsection{Per-scene Results on NDVS Datasets}
Tables~\ref{tab:comparison_24frames} and~\ref{tab:comparison_12frames} detail the per-scene NDVS results discussed in the main paper.

\input{tab/supp_ablation}
\input{tab/comparison_24frames}
\input{tab/comparison_12frames}

\subsection{Results on iPhone dataset}

\begin{figure}[t]
    \centering
    \includegraphics[width=0.5\linewidth,trim={0 77.66cm 135cm 0},clip]{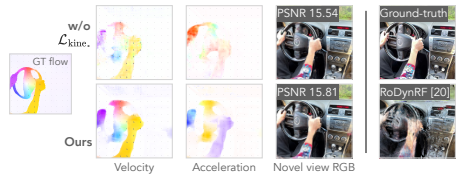}
    \caption{Comparison on ``wheel'' scene of the iPhone dataset.}
    \label{fig:iphone_qual}
\end{figure}

We provide comparison on DyCheck iPhone dataset~\cite{gao2022monocular}, realistically captured with a hand-held phone. 
In Fig.~\ref{fig:iphone_qual} and Table~\ref{tab:abl}, we ablate our kinematic loss $\mathcal{L}_\text{kine.}$ (Eq.~15). Without $\mathcal{L}_\text{kine.}$, each vel. and acc. field shows undesirable values.
Table~\ref{tab:iphone_quan} reports per-scene results on iPhone dataset, showing our competitive results. These results demonstrate that our method has the potential for broader applicability even on real-world captures.


\begin{table}[t]
\centering
\caption{Ablation on ``wheel'' scene of the iPhone dataset.}\label{tab:abl}
\begin{tabular}{r|cc|cc}
\hline
      & \multicolumn{1}{c}{T-NeRF~{\small \cite{gao2022monocular}}} & \multicolumn{1}{c|}{RoDynRF~{\small \cite{Liu_2023_CVPR}}} & \multicolumn{1}{c}{{\small w/o $\mathcal{L}_\text{kine.}$}} & \multicolumn{1}{l}{Ours} \\\hline
PSNR$\uparrow$ & 15.65                      & 15.20                        & 15.54                               & \textbf{15.75}                    \\
SSIM$\uparrow$ & 0.548                      & 0.449                       & 0.557                               & \textbf{0.558}            \\\hline       
\end{tabular}
\end{table}

\begin{table}[t]
\centering
\caption{PSNR$\uparrow$ and SSIM$\uparrow$ on each scene of the iPhone dataset.}\label{tab:iphone_quan}
\begin{tabular}{l|c|c|c|c|c|c|c|c||c|c}
\hline 
 & \multicolumn{2}{c|}{Apple} & \multicolumn{2}{c|}{Windmill} & \multicolumn{2}{c|}{Space-out} & \multicolumn{2}{c||}{Wheel} &  \multicolumn{2}{c}{Average} \\\hline
Nerfies & 
17.64 & 0.743 & 
17.38 & 0.382 & 
17.93 & 0.605 & 
13.99 & 0.455 & 
16.73 & 0.546 
\\
T-NeRF & 
17.43 & 0.728 & 
17.55 & 0.367 & 
17.71 & 0.591 & 
15.65 & 0.548 & 
17.09 & 0.558 
\\
RoDynRF & 
18.73 & 0.722 & 
16.71 & 0.321 & 
18.56 & 0.594 & 
15.20 & 0.449 & 
17.30 & 0.521 
\\\hline
Ours & 
17.84&	0.745&
17.85&	0.365&
18.31&	0.600&
15.75&	0.558&
\textbf{17.44}&	\textbf{0.567}
\\\hline
\end{tabular}

\end{table}

\section{3D and 4D Volume Structures}
Our method is configured with 
the dynamic radiance field $\mathcal{F}_\text{DY}$, the static radiance field $\mathcal{F}_\text{ST}$, and the kinematic field $\mathcal{F}_\text{K}$.
The 4D fields $\mathcal{F}_\text{DY}$ and $\mathcal{F}_\text{K}$ require $O(N^3TF)$ space, where $N$ is the spatial, $T$ is the temporal resolutions, and $F$ is the feature size for each voxel.
Similarly, the 3D field $\mathcal{F}_\text{ST}$ which does not depend on the time varible $t$, requires 
$O(N^3F)$.
As stated in the HexPlane paper~\cite{cao2023hexplane}
storing the whole volume in a naïve data structure takes about 48GB memory when $N=512$ and $T=32$.

Thus, we use a tensor-decomposition-based structure, HexPlane~\cite{cao2023hexplane}, to compress the large space into a more compact representation so that we can make the framework feasible.
Our 3D feature volume $V_\text{3D}\in \mathbb{R}^{XYZF}$ and 4D feature volume $V_\text{4D}\in \mathbb{R}^{XYZTF}$
are defined as:
\begin{align}
    V_\text{3D}=&
    \sum_{r=1}^{R_1}
    \mathbf{M}_r^{XY} \circ \mathbf{v}_r^{Z} \circ \mathbf{v}_r^{1} + 
    \sum_{r=1}^{R_2}
    \mathbf{M}_r^{XZ} \circ \mathbf{v}_r^{Y} \circ \mathbf{v}_r^{2} +
    \sum_{r=1}^{R_3}
    \mathbf{M}_r^{YZ} \circ \mathbf{v}_r^{X} \circ \mathbf{v}_r^{3},\\
    V_\text{4D}=&
    \sum_{r=1}^{R_1}
    \mathbf{M}_r^{XY} \circ \mathbf{M}_r^{ZT} \circ \mathbf{v}_r^{1} + 
    \sum_{r=1}^{R_2}
    \mathbf{M}_r^{XZ} \circ \mathbf{M}_r^{YT} \circ \mathbf{v}_r^{2} +
    \sum_{r=1}^{R_3}
    \mathbf{M}_r^{YZ} \circ \mathbf{M}_r^{XT} \circ \mathbf{v}_r^{3},
\end{align}
where $\circ$ is the outer product, $\mathbf{M}_r^{AB} \in \mathbb{R}^{AB}$ 
is a plane representing $A$-$B$ dimensions, $\mathbf{v}^{C}_r \in \mathbb{C}^{C}$ is the vector along the $C$-axis,
$\mathbf{v}_r^n \in \mathbb{R}^F$ is the vector along the $F$-axis, and $R_n$ is the number of low-rank components. Note that, we adopt the notations from the HexPlane paper~\cite{cao2023hexplane}.

When querying a coordinate $(x,y,z,t,\theta,\phi)$, we sample $F$-dimensional feature from each volume;
here, we leverage multi-resolution sampling method~\cite{Liu_2023_CVPR}
to sample features from the volumes.
We configure each radiance field with two volumes representing density and RGB, respectively.
For density fields, we employ an MLP-free design that does not depend on a ray direction $(\theta, \phi)$.
On the other hand, we use a tiny MLP which takes as inputs the sampled feature and the ray direction for the RGB. 
These designs are mostly equivalent to the default configuration of HexPlane~\cite{cao2023hexplane}.
For the kinematic field, we use a tiny MLP to produce the kinematic quantities, but without the ray direction as an input;
this makes the motion and density of particles in a scene be independent of a ray direction.

\begin{table}[ht]
\centering
\caption{
Number of low-rank components in HexPlane (or TensoRF) structures.
}\label{tab:hexplane_feat}
\begin{tabular}{l|llllll}
\hline
            & X-Y & Y-Z & X-Z & Z-T & Y-T & X-T \\\hline
Dy. Density & 16 & 4  & 4  & 16 & 4  & 4  \\
Dy. RGB     & 48 & 12 & 12 & 48 & 12 & 12 \\
St. Density & 16 & 4  & 4  & 16$^\dagger$ & 4$^\dagger$  & 4$^\dagger$  \\
St. RGB     & 48 & 12 & 12 & 48$^\dagger$ & 12$^\dagger$ & 12$^\dagger$ \\
Kinematic   & 32 & 16 & 16 & 32 & 16 & 16\\\hline
\multicolumn{7}{l}{$\dagger$: VM-factorization~\cite{chen2022tensorf}.}
\end{tabular}
\end{table}

\noindent\textbf{Plane and vector sizes.}
We report the number of low-rank components used for HexPlane (or TensoRF) structures in Table~\ref{tab:hexplane_feat}.
Note that the number of low-rank components defines the feature dimension of each plane; if the $X$-$Y$ plane has 48 components, then the shape of the plane becomes $X\times Y\times 48$.
We use \texttt{multiply} followed by \texttt{concat} for the feature fusion design since the combination shows better results than alternative designs~\cite{cao2023hexplane}.

\section{Details on Kinematic Fields}

\subsection{Coordinate Systems}
\label{sec:coord_system}
Since a volume-rendering method requires sampling color and density from a bounded space, learning a radiance field model from unbounded scenes might produce undesired artifacts in the regions outside the sampling boundaries.
Thus, there have been various methods to solve this problem~\cite{mildenhall2020nerf,zhang2020nerf++,barron2022mipnerf360}.
One common solution for unbounded forward-facing scenes is using the normalized device coordinates (NDC)~\cite{mildenhall2020nerf},
where a camera frustum with the unbounded z-axis is 
transformed into a $[-1,1]^3$ cube.
Further details regarding the conversion from NDC to world space are provided in Sec.~\ref{sec:ndc}.

Thus, we use NDC to learn the radiance fields for unbounded forward-facing scenes.
However, the physics-related elements should be computed in the world space, rather than in the NDC space.
Therefore, the kinematic quantities are represented in the world space in our method.

Let $g(\cdot)$ be a transformation which converts a coordinate in NDC to world:
\begin{equation}
    g(\mathbf{x}_\text{ndc}) = \mathbf{x}_\text{world},
\end{equation}
where $\mathbf{x}_\text{ndc}$ in NDC corresponds to $\mathbf{x}_\text{world}$ in world space.
Then, our kinematic field $\mathcal{F}_\text{K}$ takes a NDC and produce outputs in world space:
\begin{equation}
    \mathcal{F}_\text{K}(\mathbf{x}_\text{ndc}) =
    (\mathbf{v}_\text{world}, \mathbf{a}_\text{world}, \mathbf{j}_\text{world}),
\end{equation}
where $\mathbf{v}_\text{world}$, $\mathbf{a}_\text{world}$, and $\mathbf{j}_\text{world}$ are velocity, acceleration, and jerk defined in the world space.
To compute the displacement in world space, it is trivial to use Eq.~8,
resulting in a displacement in the world space.
On the other hand, in the case of computing the photometric consistency loss (Eq.~15) and the cycle consistency loss (Eq.~14), we need to compute displacement in NDC.
A displacement in NDC can be computed as:
\begin{equation}
    d_\text{ndc}(\mathbf{x}_\text{ndc}, t, \Delta t)
    = g^{-1}(g(\mathbf{x}_\text{ndc}) + d(g(\mathbf{x}_\text{ndc}), t, \Delta t)) - \mathbf{x}_\text{ndc},
\end{equation}
where $g(\cdot)$ transforms an NDC into the world space, and $g^{-1}(\cdot)$ is the inverse of $g(\cdot)$. 

\subsection{Computing Numerical Gradients}\label{sec:numgrad}
As noted in the main paper, numerical gradients can be computed as:
\begin{equation}
    \frac{\partial f(x)}{\partial x} \approx
    \frac{f(x+\epsilon) - f(x-\epsilon)}{2\epsilon},
    \label{eq:numgrad}
\end{equation}
where $\epsilon > 0$ is a small offset. 
Regarding this numerical method, we need to 
consider two factors: how to set $\epsilon$ and
how to deal with the NDC system.

\noindent \textbf{How to set $\epsilon$.}
Since we leverage an explicit feature representation for radiance and kinematic fields,
our voxel structure has a certain resolution in each phase of training.
Hence, we can leverage the voxel resolution to decide $\epsilon$~\cite{li2023neuralangelo}.

When dimensions of the radiance and kinematic fields are $X$,$Y$, $Z$, and $T$
we set $\epsilon$ to $\frac{2\lambda}{X}$ for the spatial axis and $\frac{2\lambda}{T}$ for the temporal axis, where $\lambda$ is the constant controlling the scale.
Note that we increase the spatial resolution from $22\times 22\times 22$ to $300 \times 300 \times 300$;
during the upsampling process, $\epsilon$ decreases accordingly.
In all experiments, we use $\lambda=0.2$.

\noindent \textbf{How to deal with NDC.}
Since the voxel features are 
defined in the NDC space, we 
compute initial $\epsilon$ in the NDC voxel space.
However, to compute Jacobians of the kinematic quantities we need to have $\epsilon$ in the world space to correctly compute the gradient (Eq.~\ref{eq:numgrad}).
Thus, we convert $\epsilon$
from NDC to $\epsilon_\text{world}$ in world space by:
\begin{equation}
    \epsilon_\text{world} =  0.5 \cdot \|g(x+\epsilon, y, z) - g(x-\epsilon, y, z)\|_2,
\end{equation}
where $g(\cdot)$ is the transformation from NDC to world, and we will omit
$y,z$ for concise notations in the following equation.
Having $\epsilon_\text{world}$ computed, we can rewrite Eq.~\ref{eq:numgrad} to:
\begin{equation}
    \frac{\partial f(x)}{\partial x} \approx
    \frac{
    f(g^{-1}
    (g(x) + \epsilon_\text{world})) -
    f(g^{-1}
    (g(x) - \epsilon_\text{world}))
    }
    {2\epsilon_\text{world}},
    \label{eq:numgrad2}
\end{equation}
where $g^{-1}$ is the world to NDC transformation.
Note that we can extend Eq.~\ref{eq:numgrad2} into $y$ and $z$ axis trivially.
For $t$-axis, it is not required to convert the space from NDC to world;
we can simply use Eq.~\ref{eq:numgrad} to compute derivatives with respect to time $t$.

\section{Normalized Device Coordinates}\label{sec:ndc}

In neural radiance fields, 
reconstructing scenes without considering unbounded regions can generate undesirable
artifacts.
The normalized device coordinate (NDC)
is proposed to deal with a scene unbounded
in the $z$-axis (\textit{i.e.}, infinite depth), which is effective in covering mostly forward-facing scenes.
Since a short monocular clip is usually forward-facing, NDC has been frequently utilized for 4D reconstruction from a monocular video~\cite{li2021neural,Liu_2023_CVPR,Gao_2021_ICCV}.
Similarly, we utilize NDC for the fields defined in our framework.

In this section, we describe about the 
conversion between the world coordinates and the normalized device coordinates (NDC).

\subsection{Projection Matrix}
In this description, we follow the notation given in the supplementary material of the NeRF paper~\cite{mildenhall2020nerf}.
The conversion between the two spaces is defined using the perspective projection matrix with the near plane $n$, the far plane $f$, the right bound $r$, and the left bound $l$.
The projection matrix $M$ is defined by:
\begin{equation}
    M = 
    \begin{pmatrix}
        \frac{n}{f} & 0 & 0 & 0 \\
        0 & \frac{n}{t} & 0 & 0 \\
        0 & 0 & \frac{-(f+n)}{f-n} & \frac{-2fn}{f-n} \\
        0 & 0 & -1 & 0
    \end{pmatrix}.
\end{equation}
The projection $M$ projects a homogeneous coordinate $(x,y,z,1)$ in world space
to NDC. The projected coordinates in NDC are mapped to a $[-1,1]^3$ cube.

\begin{align}
    &\begin{pmatrix}
        \frac{n}{f} & 0 & 0 & 0 \\
        0 & \frac{n}{t} & 0 & 0 \\
        0 & 0 & \frac{-(f+n)}{f-n} & \frac{-2fn}{f-n} \\
        0 & 0 & -1 & 0
    \end{pmatrix} 
    \begin{pmatrix}
    x\\y\\z\\1
    \end{pmatrix}\\
    \propto&
    \begin{pmatrix}
    \frac{n}{r} \frac{x}{-z}\\
    \frac{n}{t} \frac{y}{-z}\\
    \frac{f+n}{f-n} - \frac{2fn}{f-n}\frac{1}{-z}\\
    1
    \end{pmatrix}
    =
    \begin{pmatrix}
    x_\text{ndc}\\
    y_\text{ndc}\\
    z_\text{ndc}\\
    1
    \end{pmatrix},
\end{align}
where the coordinate $(x,y,z)$ in world space
is converted into the coordinate $(x_\text{ndc}, y_\text{ndc}, z_\text{ndc})$ in NDC space.
Note that, it is trivial to inverse the process to convert a homogeneous coordinate in NDC into world space.

\subsection{Density Transformation}
Following the notation used in Sec.~\ref{sec:numgrad}, 
let $g(\cdot)$ be the NDC to world transformation, where a point $\mathbf{x}_\text{ndc}$ in the NDC is converted to $\mathbf{x} = g(\mathbf{x}_\text{ndc})$
in world space.

In the transport loss (Eq.~12), we need to compute $\nabla \sigma$, with respect to the spatial coordinate $\mathbf{x}$ in world space.
However, if we formulate the rendering equation in NDC, the density value is referenced in NDC space, not in world space.
Thus, we need to define a density transformation function $\sigma =g_d(\sigma_\text{ndc}, \mathbf{x}_\text{ndc})$,
which transforms density $\sigma_\text{ndc}$ defined at $\mathbf{x}_\text{ndc}$ into the density $\sigma$ in world space.

Given two different coordinate systems A and B,
a small volume element $V_\text{A}$ in coordinate system A and the corresponding volume element $V_\text{B}$ in coordinate system B can be expressed as:
\begin{equation}
    V_\text{B} = |\det (J_f(\mathbf{x}_\text{A}))|V_\text{A},
\end{equation}
where $\det (J_f(\mathbf{x}_\text{A}))$ is the determinant of the Jacobian matrix $J_f$ of the coordinate transfomation $f$ at point $\mathbf{x}_\text{A}$.

Thus, we formulate density transformation $g_d$ as:
\begin{equation}
    g_d(\sigma_\text{ndc}, \mathbf{x}_\text{ndc})
    = \frac{C\sigma_\text{ndc}}{|\det (J_g(\mathbf{x}_\text{ndc}))|} = \sigma,
    \label{eq:density_transform}
\end{equation}
where $C$ is a constant.
Note that relationship between density $\rho$, mass $m$, and volume $V$ is
given by the formula $\rho = m/V$.
Assuming equal mass in two spaces, we can use Eq.~\ref{eq:density_transform}.

\section{Training Details}

\subsection{Fields}
As mentioned in the main paper, our radiance and kinematic fields consist of HexPlane~\cite{cao2023hexplane} and TensoRF~\cite{chen2022tensorf}.
The feature sizes of the planes and the vectors are specified in Table~\ref{tab:hexplane_feat}.

For RGB and kinematic outputs, we use a tiny 3-layer MLP; we use 64 dimension and 128 dimension in hidden layers for RGB and kinematic fields, respectively.
When we concatenate a direction vector, we do not use positional encoding; we use the $\ell^2$-normalization of a ray direction vector
as an input to an MLP.

We use the coarse-to-fine strategy to learn the voxels properly, where we upsample each voxel at 100; 1,000; 2,500; 5,000; 10,000; 20,000; 30,000 steps by the logarithmic voxel growth algorithm~\cite{chen2022tensorf}.
We set the initial number of voxels to range from 16,000 to 32,768 depending on the datasets, and we upscale the number to 27,000,000 ($300\times300\times300$) at the last stage of training.
For the maximum motion order, we use 3 (jerk) for the NDVS 24-frame sequences and 2 (acceleration) for the NDVS 12-frame sequences.

\subsection{Losses}
\noindent \textbf{Total variation (TV) loss.} For the total-variation (TV) loss~\cite{chen2022tensorf}, we use $0.001$ as the loss weight.
Since using TV loss for the kinematic fields often results in an equilibrium where all feature values are identical, we do not leverage the TV loss for the kinematic field.

\noindent \textbf{Photometric loss.} In the photometric loss (Eq.~14) and the cycle loss (Eq.~13), 
we need to define the maximum hop we sample $i$ within.
Like the logarithmic voxel growth algorithm~\cite{chen2022tensorf}, we increase the 
maximum hop from a two-frame gap to a three-frame gap during the training phase.

\noindent \textbf{Trajectory smoothness.}
Since higher-order kinematic quantities can generate a jerky trajectory, we regularize on the predicted higher-order terms and partial derivatives using an L2 penalty  regularizer:
\begin{equation}
    \mathcal{L}_S 
    = \|\mathbf{a}\|_2^2 + \|\mathbf{j}\|_2^2 + \|\partial \mathbf{v}/\partial t + \mathbf{v} \cdot \nabla \mathbf{v}\|_2^2 + \|\partial \mathbf{a}/\partial t + \mathbf{v} \cdot \nabla \mathbf{a}\|_2^2.
    \label{eq:ho_penalty}
\end{equation}
In addition to this loss, we start training with the first order, and gradually add higher order kinematic quantities to prevent jerky trajectories.

\noindent \textbf{Initialization w/o kinematic loss.} For stable training, we apply the kinematic loss from the 1000~th iteration; before the 1000~th, we only utilize $\mathcal{L}_\text{photo}$ and $\mathcal{L}_\text{reg}$ in the total loss $\mathcal{L}$ (Eq.~15).

\section{Discussion}
\paragraph{Limitation.}
This study introduces methods grounded in physics, utilizing kinematic quantities to describe motion and applying regularization to radiance fields. Consequently, our approach favors motion trajectories that are smooth and rigid, rather than irregular or deformable. This preference may lead to a simplification of complex motion in certain scenarios (\textit{e.g.}, jumping), where kinematic regularization might not fully capture the intricacies of real-world movement.

\paragraph{Potential negative impact.}
Our work aims to improve dynamic scene reconstruction grounded in physics.
Reconstructing dynamic scenes 
from video could potentially be exploited to create more realistic and convincing deepfakes. 
Such applications could contribute to the spread of misinformation, affecting public opinion and potentially influencing people in harmful ways.

%% file: tab/supp_ablation.tex
\begin{table}[p]
\caption{\textbf{Ablation study} on the NDVS dataset (24-frames-sparse).}\label{tab:supp_ablation}
\begin{subtable}[h]{\textwidth}
\caption{Balloon1}
\resizebox{\linewidth}{!}{
\begin{tabular}{l|cccccc|cccccc}
\hline
&\multicolumn{6}{c|}{Novel Times} & \multicolumn{6}{c}{Seen Times}   \\
&\multicolumn{3}{c}{Full} & \multicolumn{3}{c|}{Dynamic Only} &\multicolumn{3}{c}{Full} & \multicolumn{3}{c}{Dynamic Only}  \\
Methods & PSNR $\uparrow$ & SSIM $\uparrow$ & LPIPS $\downarrow$ & PSNR $\uparrow$ & SSIM $\uparrow$ & LPIPS $\downarrow$ & PSNR $\uparrow$ & SSIM $\uparrow$ & LPIPS $\downarrow$ & PSNR $\uparrow$ & SSIM $\uparrow$ & LPIPS $\downarrow$\\\hline
NSFF~\cite{li2021neural} & 
13.18&	0.146&	0.603&	14.21&	0.201&	0.574&12.49&	0.153&	0.499&	13.16&	0.202&	0.487  \\
HexPlane~\cite{cao2023hexplane} & 
19.84&	0.668&	0.145&	18.12&	0.419&	0.251& 18.93&	0.649&	0.137&	17.74&	0.429&	0.202  \\\hline
\textbf{Ours} & 
\textbf{25.67}&	\textbf{0.865}&	\textbf{0.064}&	\textbf{21.16}&	\textbf{0.625}&	\textbf{0.170}& \textbf{25.80}&	\textbf{0.867}&	\textbf{0.050}&	\textbf{21.35}&	\textbf{0.632}&	\textbf{0.121}\\\hline
w/o Higher order motion &
24.36&	0.842&	0.077&	19.93&	0.556&	0.200& 24.63&	0.846&	0.061&	20.22&	0.572&	0.145 \\
w/o $\mathcal{L}_\text{kinematic}$ & 
24.96&	0.853&	0.068&	20.15&	0.568&	0.185 & 25.30&	0.858&	0.052&	20.65&	0.594&	0.129\\
w/o Kinematic Integrity & 
25.46&	0.862&	0.066&	20.78&	0.607&	0.179 & 25.57&	0.865&	0.051&	21.02&	0.620&	0.126\\
w/o Rigidity &  
25.11&	0.856&	0.066&	20.40&	0.582&	0.181 & 25.43&	0.860&	0.051&	20.82&	0.603&	0.123 \\\hline
\end{tabular}
}
\end{subtable}
\begin{subtable}[h]{\textwidth}
\caption{DynamicFace}
\resizebox{\linewidth}{!}{
\begin{tabular}{l|cccccc|cccccc}
\hline
&\multicolumn{6}{c|}{Novel Times} & \multicolumn{6}{c}{Seen Times}   \\
&\multicolumn{3}{c}{Full} & \multicolumn{3}{c|}{Dynamic Only} &\multicolumn{3}{c}{Full} & \multicolumn{3}{c}{Dynamic Only}  \\
Methods & PSNR $\uparrow$ & SSIM $\uparrow$ & LPIPS $\downarrow$ & PSNR $\uparrow$ & SSIM $\uparrow$ & LPIPS $\downarrow$ & PSNR $\uparrow$ & SSIM $\uparrow$ & LPIPS $\downarrow$ & PSNR $\uparrow$ & SSIM $\uparrow$ & LPIPS $\downarrow$\\\hline
NSFF~\cite{li2021neural} & 
10.95&	0.131&	0.539&	8.98&	0.064&	0.628&10.77	&0.120&	0.545&	8.88&	0.055&	0.636   \\
HexPlane~\cite{cao2023hexplane} & 
12.11&	0.265&	0.471&	14.38&	0.431&	0.397&11.39&	0.218&	0.491&	13.74&	0.442&	0.395    \\\hline
\textbf{Ours} & 
26.47&	\textbf{0.896}&	0.051&	\textbf{25.22}&	\textbf{0.893}&	\textbf{0.028} & 25.20&	\textbf{0.881}&	0.060&	\textbf{24.74}&	\textbf{0.887}&	\textbf{0.031}	\\\hline
w/o Higher order motion &
\textbf{26.49}&	\textbf{0.896}&	0.051&	25.03&	0.882&	0.031&\textbf{25.22}&	\textbf{0.881}&	0.059&	24.47&	0.874&	0.034  \\
w/o $\mathcal{L}_\text{kinematic}$ & 
26.24&	0.895&	0.052&	24.34&	0.867&	0.034&	24.98&	0.880&	0.060&	23.87&	0.859&	0.037\\
w/o Kinematic Integrity & 
26.24&	0.898&	\textbf{0.050}&	24.65&	0.871&	0.032&	25.02&	0.884&	\textbf{0.057}&	24.08&	0.861&	0.036\\
w/o Rigidity &  
26.20&	0.897&	0.051&	24.73&	0.878&	0.032&	24.95&	0.883&	0.059&	24.19&	0.870&	0.035 \\\hline
\end{tabular}
}
\end{subtable}

\begin{subtable}[h]{\textwidth}
\caption{Playgruond}
\resizebox{\linewidth}{!}{
\begin{tabular}{l|cccccc|cccccc}
\hline
&\multicolumn{6}{c|}{Novel Times} & \multicolumn{6}{c}{Seen Times}   \\
&\multicolumn{3}{c}{Full} & \multicolumn{3}{c|}{Dynamic Only} &\multicolumn{3}{c}{Full} & \multicolumn{3}{c}{Dynamic Only}  \\
Methods & PSNR $\uparrow$ & SSIM $\uparrow$ & LPIPS $\downarrow$ & PSNR $\uparrow$ & SSIM $\uparrow$ & LPIPS $\downarrow$ & PSNR $\uparrow$ & SSIM $\uparrow$ & LPIPS $\downarrow$ & PSNR $\uparrow$ & SSIM $\uparrow$ & LPIPS $\downarrow$\\\hline
NSFF~\cite{li2021neural} & 
12.54&	0.113&	0.581&	11.58&	0.072&	0.629&12.15	&0.117&	0.515&	11.41&	0.067&	0.530   \\
HexPlane~\cite{cao2023hexplane} & 
14.79&	0.286&	0.341&	13.40&	0.191&	0.379&13.80&	0.230&	0.341&	12.66&	0.163&	0.360   \\\hline
\textbf{Ours} & 
\textbf{24.64}&	\textbf{0.832}&	\textbf{0.068}&	\textbf{18.24}&	\textbf{0.621}&	\textbf{0.141}&	\textbf{23.82}&	\textbf{0.811}&	\textbf{0.074}&	\textbf{18.14}&	\textbf{0.621}&	\textbf{0.133}\\\hline
w/o Higher order motion &
24.42&	0.829&	\textbf{0.068}&	17.63&	0.576&	0.149&23.58&	0.808&	0.075&	17.45&	0.569&	0.144  \\
w/o $\mathcal{L}_\text{kinematic}$ & 
24.38&	0.830&	\textbf{0.068}&	17.61&	0.595&	0.153&23.57&	0.809&	0.075&	17.49&	0.596&	0.147\\
w/o Kinematic Integrity & 
24.52&	0.830&	0.069&	17.93&	0.605&	0.150&23.69&	0.809&	0.075&	17.81&	0.605&	0.140\\
w/o Rigidity &  
24.40&	0.830&	\textbf{0.068}&	17.64&	0.597&	0.146&23.62&	0.810&	0.075&	17.65&	0.603&	0.137 \\\hline
\end{tabular}
}
\end{subtable}

\begin{subtable}[h]{\textwidth}
\caption{Skating}
\resizebox{\linewidth}{!}{
\begin{tabular}{l|cccccc|cccccc}
\hline
&\multicolumn{6}{c|}{Novel Times} & \multicolumn{6}{c}{Seen Times}   \\
&\multicolumn{3}{c}{Full} & \multicolumn{3}{c|}{Dynamic Only} &\multicolumn{3}{c}{Full} & \multicolumn{3}{c}{Dynamic Only}  \\
Methods & PSNR $\uparrow$ & SSIM $\uparrow$ & LPIPS $\downarrow$ & PSNR $\uparrow$ & SSIM $\uparrow$ & LPIPS $\downarrow$ & PSNR $\uparrow$ & SSIM $\uparrow$ & LPIPS $\downarrow$ & PSNR $\uparrow$ & SSIM $\uparrow$ & LPIPS $\downarrow$\\\hline
NSFF~\cite{li2021neural} & 
18.18&	0.411&	0.418&	12.41&	0.136&	0.608 &17.20&	0.397&	0.364&	11.75&	0.124&	0.513  \\
HexPlane~\cite{cao2023hexplane} & 
22.38&	0.783&	0.151&	12.43&	0.145&	0.352&21.77&	0.769&	0.154&	12.12&	0.140&	0.357    \\\hline
\textbf{Ours} & 
\textbf{30.99}&	\textbf{0.941}&	0.033&	\textbf{18.60}&	\textbf{0.563}&	\textbf{0.165}&\textbf{31.91}&	\textbf{0.943}&	0.029&	\textbf{20.57}&	\textbf{0.671}&	\textbf{0.107}\\\hline
w/o Higher order motion &
30.89&	0.940&	\textbf{0.032}&	18.45&	0.536&	0.161&31.36&	0.940&	\textbf{0.028}&	19.70&	0.610&	0.109  \\
w/o $\mathcal{L}_\text{kinematic}$ & 
5.88&	0.463&	0.698&	3.73&	0.138&	0.857&5.88&	0.463&	0.698&	3.74&	0.141&	0.859\\
w/o Kinematic Integrity & 
26.04&	0.916&	0.059&	14.59&	0.316&	0.267&26.50&	0.918&	0.055&	14.94&	0.372&	0.228\\
w/o Rigidity &  
29.72&	0.935&	0.036&	16.98&	0.425&	0.198&30.05&	0.935&	0.032&	17.78&	0.492&	0.150 \\\hline
\end{tabular}
}
\end{subtable}

\end{table}

%% file: tab/comparison_24frames.tex
\begin{table*}[!ht]
\caption{\textbf{Quantitative comparison} on NDVS (24-frames). Numbers that surpass their counterparts by a margin greater than 5\% are highlighted.}
\label{tab:comparison_24frames}
\resizebox{\linewidth}{!}{
\begin{tabular}{lc|ccc|ccc|ccc|ccc}
\hline 
\multicolumn{1}{l}{method} & & \multicolumn{3}{c|}{Playground} & \multicolumn{3}{c|}{Balloon1} & \multicolumn{3}{c|}{Balloon2} & \multicolumn{3}{c}{Umbrella} \\
& & PSNR $\uparrow$ & SSIM $\uparrow$ & LPIPS $\downarrow$ & PSNR $\uparrow$ & SSIM $\uparrow$ & LPIPS $\downarrow$ & PSNR $\uparrow$ & SSIM $\uparrow$ & LPIPS $\downarrow$ &  PSNR $\uparrow$ & SSIM $\uparrow$ & LPIPS $\downarrow$\\
\hline 
NSFF~\cite{li2021neural} & full & 24.80 & 0.865 & 0.056 & 24.14 & 0.848 & 0.064 & 29.34 & 0.918 & 0.034 & 24.46 & 0.787 & \textbf{0.087 }\\
(train: 96~h) & dyn. & 19.12 & 0.671 & 0.120 & 18.29 & 0.475 & 0.183 & 21.28 & 0.670 & 0.112 & 17.00 & 0.444 & 0.155 \\
\hline 
Ours  & full         &\textbf{26.65}&	\textbf{0.911}&	\textbf{0.042}&	\textbf{27.26}&	\textbf{0.893}&	\textbf{0.043}&	29.50&	0.916&	0.035&	\textbf{26.24}&	\textbf{0.792}&	0.098 \\
(train: 7~h)  & dyn. &18.75&	0.658&	0.122&	\textbf{21.85}&	\textbf{0.654}&	\textbf{0.117}&	22.23&	0.692&	\textbf{0.104}&	\textbf{22.25}&	\textbf{0.695}&	\textbf{0.097} \\
\hline
\end{tabular}
}
\resizebox{\linewidth}{!}{
\begin{tabular}{lc|ccc|ccc|ccc|ccc}
\hline 
\multicolumn{1}{l}{method} & &  \multicolumn{3}{c|}{Jumping} & \multicolumn{3}{c|}{DynamicFace} & \multicolumn{3}{c|}{Skating} & \multicolumn{3}{c}{Truck} \\
& & PSNR $\uparrow$ & SSIM $\uparrow$ & LPIPS $\downarrow$ & PSNR $\uparrow$ & SSIM $\uparrow$ & LPIPS $\downarrow$ & PSNR $\uparrow$ & SSIM $\uparrow$ & LPIPS $\downarrow$ & PSNR $\uparrow$ & SSIM $\uparrow$ & LPIPS $\downarrow$ \\
\hline 
NSFF~\cite{li2021neural} & full & 26.84 & 0.893 & 0.053 & 26.993 & 0.947 & 0.022 & \textbf{35.10} & \textbf{0.969} & \textbf{0.018} & 32.70 & 	0.940	& \textbf{0.022} \\
(train: 96~h) & dyn. & 19.60 & 0.616 & 0.130 & 25.311 & 0.886 & 0.028 & \textbf{22.28} & \textbf{0.724} & 0.109 & 27.86&	0.840&	\textbf{0.054}\\
\hline 
Ours & full & 26.55&	0.895&	\textbf{0.049}&	\textbf{29.15}&	0.967&	\textbf{0.011}&	32.33&	0.959&	0.024&	32.57&	0.930&	0.033\\
(train: 7~h)  & dyn. & 19.16&	0.603&	\textbf{0.114}&	25.68&	0.901&   \textbf{0.018}&	19.58&	0.626&	0.108&	27.54&	0.854&	0.068\\
\hline
\end{tabular}
}
\end{table*}

%% file: tab/comparison_12frames.tex
\begin{table*}[!ht]
\caption{\textbf{Quantitative comparison} on NDVS (12-frames).}
\label{tab:comparison_12frames}
\centering
\resizebox{\linewidth}{!}{
\begin{tabular}{l|c|c|c|c|c|c|c||c}
\hline 
PSNR$\uparrow$/LPIPS$\downarrow$ & Jumping & Skating & Truck & Umbrella & Balloon1 & Balloon2 & Playground & Average \\\hline
NeRF~\cite{mildenhall2020nerf}* & 20.99/0.305 & 23.67/0.311 & 22.73/0.229& 21.29/0.440& 19.82/0.205& 24.37/0.098& 21.07/0.165 & 21.99/0.250 \\
NSFF~\cite{li2021neural}* & 24.65/0.151&29.29/0.129&25.96/0.167&22.97/0.295&21.96/0.215&24.27/0.222&21.22/0.212&24.33/0.199 \\
DynamicNeRF~\cite{Gao_2021_ICCV}* & 24.68/0.090&32.66/0.035&28.56/0.082&23.26/0.137&22.36/0.104&27.06/0.049&24.15/0.080&\underline{26.10}/0.082 \\
HyperNeRF~\cite{park2021hypernerf}* & 18.34/0.302&21.97/0.183&20.61/0.205&18.59/0.443&13.96/0.530&16.57/0.411&13.17/0.495&17.60/0.367 \\
RoDynRF~\cite{Liu_2023_CVPR}* & 25.66/ 0.071&28.68/0.040&29.13/0.063&24.26/0.089&22.37/0.103&26.19/0.054&24.96/0.048&25.89/\textbf{0.065} \\\hline
Ours & 
23.74/0.102&
31.89/0.035&
28.34/0.074&
25.46/0.098&
23.72/0.079&
26.49/0.055&
24.64/0.053&
\textbf{26.33}/\underline{0.071}\\\hline
\multicolumn{9}{l}{*Results reported in \cite{Liu_2023_CVPR}}
\end{tabular}
}

\label{tab:rodynrf}
\end{table*}